\documentclass[journal]{IEEEtran}

\usepackage[numbers,sort&compress]{natbib} 
\usepackage{float}

\usepackage{multirow}
\usepackage{threeparttable}

\usepackage[ruled,linesnumbered]{algorithm2e} 
\usepackage{amsmath}
\usepackage{amsfonts}
\usepackage{amssymb}  

\usepackage{graphicx}
\graphicspath{{images/}}                     

\usepackage{hyperref}

\usepackage[utf8]{inputenc}

\usepackage{pifont}

\usepackage{booktabs}

\usepackage{soul}
\usepackage{color, xcolor}
\soulregister\cite7
\soulregister\citep7
\soulregister\citet7 
\soulregister\ref7 
\soulregister\pageref7 
\usepackage{xcolor}
\usepackage{colortbl}

\definecolor{myyellow}{RGB}{255,255,153} 
\definecolor{myred}{RGB}{255,153,153}   
\definecolor{myorg}{RGB}{255,165,0} 
\definecolor{mywhite}{RGB}{255,255,255}

\usepackage{makecell}

\usepackage{dcolumn}
\newcolumntype{C}{>{\arraybackslash}c} 
\newcolumntype{D}{D{/}{/}{5}} 

\begin{document}
%


\title{FoBa: A Foreground-Background co-Guided Method and New Benchmark for Remote Sensing Semantic Change Detection}

\author{Haotian Zhang$^{1}$, Han Guo$^{1}$, Keyan Chen$^{1}$, Hao Chen$^{2}$, Zhengxia Zou$^{1}$, and Zhenwei Shi$^{1, \star}$ 
\\
\vspace{6pt}
Beihang University$^1$, Shanghai Artificial Intelligence Laboratory$^2$

}
\date{September. 2025}

\maketitle
\begin{abstract}


Despite the remarkable progress achieved in remote sensing semantic change detection (SCD), two major challenges remain. At the data level, existing SCD datasets suffer from limited change categories, insufficient change types, and a lack of fine-grained class definitions, making them inadequate to fully support practical applications. At the methodological level, most current approaches underutilize change information, typically treating it as a post-processing step to enhance spatial consistency, which constrains further improvements in model performance. To address these issues, we construct a new benchmark for remote sensing SCD, LevirSCD. Focused on the Beijing area, the dataset covers 16 change categories and 210 specific change types, with more fine-grained class definitions (e.g., roads are divided into unpaved and paved roads). Furthermore, we propose a foreground-background co-guided SCD (FoBa) method, which leverages foregrounds that focus on regions of interest and backgrounds enriched with contextual information to guide the model collaboratively, thereby alleviating semantic ambiguity while enhancing its ability to detect subtle changes. Considering the requirements of bi-temporal interaction and spatial consistency in SCD, we introduce a Gated Interaction Fusion (GIF) module along with a simple consistency loss to further enhance the model’s detection performance. Extensive experiments on three datasets (SECOND, JL1, and the proposed LevirSCD) demonstrate that FoBa achieves competitive results compared to current SOTA methods, with improvements of 1.48\%, 3.61\%, and 2.81\% in the SeK metric, respectively. Our code and dataset are available at https://github.com/zmoka-zht/FoBa.
\end{abstract}


\IEEEpeerreviewmaketitle


\section{Introduction}
\label{sec:intro}

\IEEEPARstart{C}{hange} detection (CD) is a fundamental task in the field of earth observation, aiming to identify land cover changes in specific regions (including both the changed areas and the types of changes) by utilizing multi-temporal remote sensing images. It plays a crucial role in advancing our understanding of the interactions between human activities and the natural environment. As a result, the task has attracted sustained attention from the remote sensing (RS) research community and has been widely applied in several key areas, including land resource management {\cite{Li2016SLC-CD, AMGIC, lu2004change}}, disaster assessment {\cite{zheng2021building, chen2024changemamba}}, urban expansion monitoring {\cite{coppinp2004digitalchangedetection, zhang2024bifa, CDMamba}}, and geographic information system (GIS) updating {\cite{guo2021deep, LCCDReview}}.

According to the form of outputs produced by change detectors, CD tasks can be roughly divided into two categories: binary change detection (BCD) and semantic change detection (SCD). BCD aims to locate the regions of interest where changes have occurred, providing coarse-grained information about the changes without specifying their types. However, in many applications, there is a demand not only for identifying where changes occur but also for understanding what specific changes have taken place. For this reason, the SCD task has been proposed. Unlike BCD, which adopts a single-label representation, SCD employs a "from-to" labeling scheme (assigning semantic categories to each image separately) enabling a more effective characterization of key information such as semantic transitions throughout the change process.

Early SCD methods primarily relied on low-level image features such as texture, spectral information, and color differences to extract change regions, followed by the use of classifiers such as random forest or support vector machines (SVM) to distinguish change categories {\cite{bovolo2006theoretical, chen2003land}}. Another category of methods is the post-classification comparison (PCC) {\cite{huang2010sampling, hu2013seasonal, wu2017post, bruzzone2004detection}}, which first performs pixel-wise classification on bi-temporal images independently, and then derives semantic change detection results by comparing the resulting classification maps. In addition, some researchers have employed object-based image analysis techniques to reduce the boundary-related errors commonly associated with the PCC {\cite{hussain2013change, liu2021change, wang2018object}}. However, these traditional methods generally rely on handcrafted or selectively chosen features, making them sensitive to sensor characteristics and illumination variations, which significantly limits their detection performance.

With the advancement of deep learning techniques and the proposal of some high-quality SCD datasets (e.g., SECOND {\cite{yang2021asymmetric}}, MSSCD {\cite{he2023spatial}}, Landsat {\cite{landsat}}, JL1, etc.), traditional SCD methods have been gradually replaced by deep learning-based approaches. Existing deep learning-based SCD methods can be broadly categorized into single-branch, dual-branch, and multi-task methods according to the form of their detector outputs. Single-branch methods (also referred to as direct SCD approaches) abandon the "from-to" representation and instead define each type of change as an independent category, similar to the concept used in semantic segmentation {\cite{daudt2019multitask, peng2019end, ren2021improved, wang2025change}}. Although such methods are more straightforward, they tend to aggravates the issue of class imbalance, especially in scenarios involving a large number of change types. The dual-branch methods employ two separate semantic segmentation networks to independently predict the semantic categories of each single-temporal image, followed by a comparison to detect changes {\cite{xia2022deep, mou2018learning}}, or alternatively, directly predict "from-to" change maps through different branches {\cite{peng2021scdnet, tang2024clearscd}}. Although dual-branch SCD methods avoid the severe class imbalance issues encountered by single-branch approaches, they lack constraints on the predicted change results, making it difficult to ensure spatial consistency (i.e., maintaining consistent shapes of the change regions). The multi-task SCD methods extend dual-branch approaches by incorporating the BCD task, enabling simultaneous prediction of both change categories and change masks. The mask information is then integrated into the SCD outputs to enhance spatial consistency {\cite{yang2021asymmetric, tang2024clearscd, BiSRNet, TED, DEFO, cdsc, chen2024changemamba, zhou2024late}}. Owing to their excellent performance, the multi-task approaches have become the mainstream paradigm in current SCD research.

Although the aforementioned datasets and methods has contributed to the advancement of the SCD field to some extent, several challenges remain. 1) At the data level, existing SCD datasets constructed from real remote sensing images include a relatively small number of change categories (for example, the SECOND dataset includes 6 change classes, the JL1 dataset 5 change classes, and the Landsat dataset only 4 change classes). Moreover, the category granularity is relatively coarse, making it difficult to meet the requirements of diverse and complex real-world scenarios. 2) At the method level, current mainstream multi-task SCD methods still underutilize change region information, usually incorporating it only as a post-processing step to enhance spatial consistency. This ignores its potential guiding role at intermediate feature levels, thereby limiting the effective modeling of precise change information.

To address the above challenges, we propose the LevirSCD semantic change detection dataset at the data level, and a foreground-background co-guided semantic change detection (FoBa) method at the methodological level. Specifically, the LevirSCD dataset consists of 3225 image pairs, covering 16 change categories and providing finer-grained annotations (for example, roads are further divided into unpaved road and paved road), covering 210 types of changes. The core idea of the proposed FoBa is to jointly leverage guidance from both change regions (foreground) and unchanged regions (background) for modeling, thereby fully exploiting the information contained in the change areas. Different from the methods that rely solely on change-region guidance {\cite{change_guided}}, the proposed approach incorporates background information with sufficient contextual content to alleviate semantic ambiguity. Additionally, it helps mitigate the tendency to overemphasize prominent changes, thereby improving the model's sensitivity to subtle changes. Furthermore, we propose a Gated Interaction Fusion (GIF) module to enhance the interaction between bi-temporal features and a simple consistency loss is introduced to constrain the unchanged regions, thereby improving the overall detection performance of the model. 

In summary, the main contributions of this paper can be summarized as follows:
\begin{itemize}
\item Propose a semantic change detection dataset LevirSCD based on the real-word remote sensing images. This dataset features a rich variety of change categories, fine-grained annotations, and diverse change types, and is expected to provide valuable data support for research in the SCD field.

\item Propose a novel foreground-background co-guided semantic change detection (FoBa) method that leverages a joint guidance mechanism from both change regions and unchanged regions to fully exploit change information, thereby achieving more accurate SCD. 

\item Quantitative and qualitative experiments conducted on three datasets demonstrate that the proposed FoBa achieves superior performance.
\end{itemize}

The remainder of this paper is organized as follows: Section \ref{sec:relatedwork} reviews related work. Section \ref{sec:levirscd} introduces the constructed LevirSCD dataset. Section \ref{sec:method} provides a detailed description of the proposed FoBa. Some experimental results are reported in Section \ref{sec:experiment}. And the conclusion is made in Section \ref{sec:conclusion}.

\begin{table*}
    \centering
    \caption{Commonly used open-source SCD  datasets and their basic information.}
    \resizebox{0.9\textwidth}{!}{
    \begin{tabular}{l|c|c|c|c|c|c|c}
    \toprule
    Dataset & Change cat. & Nums. & Fine granularity & Object-level annotation & Time span & Size & Resolution \\
    \midrule
    Hanyang$_{2016}$    & 7 & 1 & $\times$ & $\times$ & 2002-2009 & 7200$\times$6000 & 1\\
    HRSCD$_{2019}$     & 5 & 291 & $\times$ & $\times$ & 2006-2012 & 10000$\times$10000 & 0.5\\
    Hi-UCD$_{2020}$    & 9 & 1293 & $\times$ & $\checkmark$ & 2017-2019 & 1024$\times$1024 & 0.1\\
    Landsat$_{2022}$   & 9 & 8468 & $\times$ & $\times$ & 1990-2020 & 416$\times$416 & 30 \\
    SECOND$_{2022}$    & 6 & 4662 & $\times$ & $\checkmark$ & - & 512$\times$512 & 0.5-3 \\
    DynamicEarthNet$_{2022}$ & 7 & 54750 & $\times$ &$\checkmark$ & 2018-2019 & 1024$\times$1024 & 3 \\
    CNAM-CD$_{2023}$   & 5 & 2508 & $\times$ &  $\times$ & 2013-2022 & 512$\times$512 & 0.5 \\
    WUSU$_{2023}$ & 11 & 2 & $\checkmark$ & $\checkmark$  & 2015-2018 & 6358$\times$6382/7025$\times$5500 & 1 \\
    JL1$_{2023}$       & 5 & 6000 & $\times$ & $\times$ & - & 256$\times$256 & 0.75 \\
    \midrule
    LevirSCD & 16 & 3225 & $\checkmark$ & $\checkmark$ & 2010-2019& 256$\times$256 & 1-2 \\
    \bottomrule
    \end{tabular}
    }
    \label{tab:scd_datasets}
\end{table*}


\section{Related Work} \label{relatedwork}
\label{sec:relatedwork}

\subsection{SCD dataset}

CD is a critical task in earth observation and has attracted significant attention from the research community. Several datasets have been developed to support CD tasks, such as WHU-CD {\cite{WHU-CD}}, LEVIR-CD {\cite{LEVIR-CD}}, and SYSU-CD {\cite{SYSU-CD}}. However, these datasets primarily focus on specific objects (e.g., buildings) or change regions, and lack comprehensive semantic information about diverse change types. 

To address the above issues, several researchers have focused on constructing SCD datasets {\cite{he2023spatial, landsat, yang2021asymmetric, Mts-WH-2016, hrscd4, Hi-UCD-2020, DynamicEarthNet-2022, CNAM-CD-2023, WUSU-2023}}. The commonly used open-source SCD datasets are summarized in Table \ref{tab:scd_datasets}. Wu et al. proposed the Hanyang dataset {\cite{Mts-WH-2016}} in 2016, which was collected using the IKONOS sensor over a temporal span from 2002 to 2009. The image size is 7200$\times$6000, and a spatial resolution of 1m/pixel. The image was partitioned into 150$\times$150 pixel patches, each assigned a semantic label to construct a scene-level SCD dataset. The dataset encompasses seven change categories, including common land cover types such as water and farmland. In contrast to the scene-level annotations used in the aforementioned datasets, the HRSCD dataset proposed by Daudt et al. {\cite{hrscd4}} adopts more fine-grained pixel-level annotations. It consists of 291 pairs of remote sensing images, each with a size of 10000$\times$10000 pixels, covering five types of land cover changes. And, the spatial resolution is 0.5 m/pixel. Subsequently, a variety of SCD has emerged in rapid succession. Examples include high-resolution datasets focusing on urban scenes such as Hi-UCD {\cite{Hi-UCD-2020}}, SECOND {\cite{yang2021asymmetric}}, DynamicEarthNet {\cite{DynamicEarthNet-2022}}, CNAM-CD {\cite{CNAM-CD-2023}}, and WUSU {\cite{WUSU-2023}}. Meanwhile, lower-resolution datasets such as Landsat {\cite{landsat}} have also been introduced, along with the JL1 dataset designed for competition tasks.

In summary, although several publicly available SCD datasets have been released, the range of change categories they cover remains limited. Compared to the WUSU dataset, which currently includes the largest number of change categories, the proposed LevirSCD dataset achieves an approximate 45\% increase in change category diversity. Moreover, existing public datasets rarely offer both fine-grained and object-level annotations. The introduction of the LevirSCD dataset is expected to further enrich research in this aspect.

\subsection{BCD}

BCD primarily focuses on identifying change regions in bi-temporal images. In recent years, the rapid advancement of deep learning techniques has significantly promoted the development of the BCD field. In early studies, Daudt et al. {\cite{daudt2018FC-EF}} proposed a fully convolutional network-based method, FC-EF, which performs BCD by concatenating bi-temporal images along the channel dimension. They also introduced two Siamese CNN-based variants, FC-Siam-Conc and FC-Siam-Diff, for processing the input images. Subsequently, a series of CNN-based methods have been proposed. Fang et al. {\cite{fang2021snunet}} established deep interactions between bi-temporal images using a densely connected CNN architecture. Some studies further introduced deep supervision mechanisms to guide the difference features extracted from various CNN stages, thereby enabling multi-level fine-grained detection {\cite{zhang2020DSIFN, shi2021deeply}}. To enhance the discriminability of features, Lei et al. {\cite{lei2021difference}} proposed the difference enhancement network, which effectively learns the differential representations between foreground and background regions. Jiang et al. {\cite{jiang2022wricnet}} weighted multi-scale network that adaptively assigns weights to features at different scales to accommodate change regions of varying sizes. Huang et al. {\cite{huang2021multiple}} employed selective convolutional kernels and multiple attention mechanisms to achieve selective fusion of bi-temporal features. In addition, some studies have incorporated auxiliary tasks to enhance BCD performance. For example, Liu et al. {\cite{liu2021super}} integrated a super-resolution task into the change detection model to mitigate cumulative errors arising from bi-temporal images with differing resolutions.

However, due to the inherent limitations of CNNs in global context modeling, Transformer-based approaches have been proposed in recent years to address this issue. Chen et al. {\cite{BIT}} proposed the BIT, which represents bi-temporal features as visual-semantic tokens and employs a Transformer to achieve efficient global context modeling. Song et al. {\cite{song2022ACABFNet}} utilized a Transformer-based axial cross-attention mechanism to capture global relationships. Feng et al. {\cite{ICIFNet}} employed the Transformer to model both intra-scale cross-interaction and inter-scale feature fusion of bi-temporal images. Building upon this, they further proposed DMINet, which uses channel-wise concatenated bi-temporal features as shared queries to learn spatiotemporal relationships between images {\cite{DMINet}}. Zhang et al. {\cite{zhang2024bifa}} introduced a Transformer-based bi-temporal feature extraction method to mitigate false detections caused by irrelevant changes such as illumination and seasonal variations.

Although Transformer-based methods achieve promising performance in the tasks, their quadratic computational complexity results in significant computational overhead. Consequently, some researchers have begun to explore BCD approaches based on Mamba, which offers linear time complexity. Zhao et al. {\cite{RS-Mamba}} proposed RS-Mamba by introducing a four-diagonal scanning mechanism to extend VMamba {\cite{liu2024vmamba}}, applying it to both image segmentation and BCD tasks. Around the same time, Chen et al. {\cite{chen2024changemamba}} introduced ChangeMamba, which captures spatiotemporal dependencies by rearranging bi-temporal features. Zhang et al. {\cite{CDMamba}} proposed CDMamba to address the deficiency of existing Mamba-based methods in handling dense prediction tasks due to the lack of local cues. Since then, a series of Mamba-based methods has emerged. For example, Zhou et al. {\cite{zhou2025sprmamba}} introduced a windowing operation and cross-window interaction mechanism to alleviate the insufficient model response to subtle changes. Wu et al. {\cite{wu2025xlstm}} proposed a temporal-aware interaction module based on the xLSTM network to adaptively learn the spatiotemporal relationships. Fu et al. {\cite{fu2025beyond}} applied the concept of histogram equalization to normalize bi-temporal features, aiming to address the style discrepancies between bi-temporal images.

Distinct from the aforementioned methods for BCD, our work focuses on SCD, which involves richer semantic information and better aligns with real-world scenarios

\begin{figure*}
    \centering
    \includegraphics[width=0.98\textwidth]{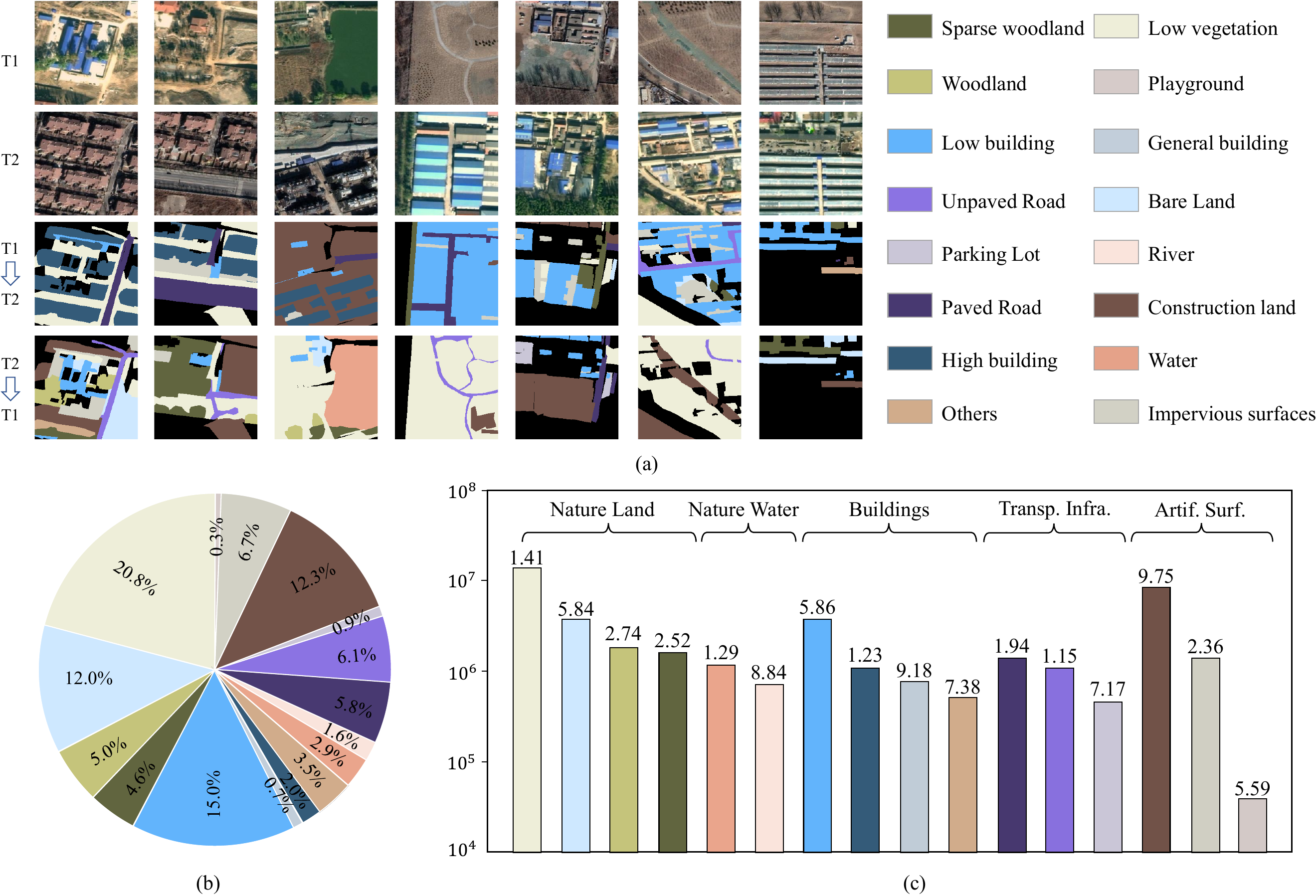}
    \caption{Sample visualization and statistical analysis of the LevirSCD dataset. (a) Some representative samples from the LevirSCD dataset. (b) The occurrence frequency of samples for different classes. (c) The number of pixels for each class. Note, the different colors correspond to the classes on the right side of (a).}
    \label{fig:levirscd_vis_static}
\end{figure*}

\subsection{SCD}

Unlike BCD, which focuses solely on identifying changed regions, SCD involves richer semantic information. Due to its closer alignment with real-world scenarios, it has gradually attracted increasing attention from the research community. Deep learning-based SCD methods can be roughly categorized into three types based on the output of their detectors: single-branch, dual-branch, and multi-task approaches. Single-branch methods treat each type of change as a distinct class, resembling conventional semantic segmentation tasks. As a pioneer, Daudt et al. {\cite{daudt2019multitask}} employed a fully convolutional network (FCN) to perform SCD. Ren et al. {\cite{ren2021improved}} proposed an improved U-Net architecture that integrates asymmetric convolution to further enhance feature extraction capability. Addressing the limitation that existing SCD methods primarily focus on visual cues while overlooking language information, Wang et al. {\cite{wang2025change}} introduced a change knowledge-guided approach that incorporates language cues related to change knowledge, thereby enhancing semantic understanding and the representation of fine-grained change details. Although single-branch methods are more intuitive, they often suffer from severe class imbalance, which adversely affects model training. As a result, their application remains relatively limited.

Dual-branch methods employ two semantic segmentation networks to independently predict the semantic categories of each temporal image. The two results are then compared to generate the final semantic change map. Peng et al. {\cite{peng2021scdnet}} proposed a Siamese UNet-based approach that integrates multi-scale atrous convolution modules to capture multi-scale information, along with a deep supervision strategy to enhance model performance. Xia et al. {\cite{xia2022deep}} introduced a deep Siamese post-classification fusion method that mitigates error accumulation in SCD by incorporating a temporal correlation and soft fusion mechanisms. Zhang et al. {\cite{he2023spatial}} proposed a semi-supervised contrastive learning framework, which leverages contrastive learning with an adaptive sampling strategy to alleviate the class imbalance problem. Although two-branch SCD methods alleviate the severe class imbalance encountered by single-branch approaches to some extent, they lack constraints on the predicted results, making it difficult to ensure spatial consistency in the detected change regions.

Multi-task methods build upon two-branch approaches by integrating BCD tasks, simultaneously predicting change category labels and change masks. The mask information is then incorporated into the SCD predictions to enhance spatial consistency. Yang et al. {\cite{yang2021asymmetric}} proposed the asymmetric siamese network, which identifies semantic changes by capturing features through structurally different modules. Wang et al. {\cite{wang2020coarse}} introduced a coarse-to-fine framework that determines the final change categories by applying majority voting over CNN-based predictions. Zheng et al. {\cite{zheng2022changemask}} developed ChangeMask, which employs a 3D convolution-based temporal-symmetric transformer module to learn highly discriminative and temporally symmetric feature representations. Ding et al. {\cite{BiSRNet}} proposed the BiSRNet, incorporating siamese semantic reasoning and cross-temporal semantic reasoning modules to model temporal correlations. Additionally, they introduced a semantic consistency loss based on contrastive learning to enhance semantic coherence. Building upon this, they further proposed the semantic change transformer, which explicitly models the "from-to" semantic transitions between bi-temporal images {\cite{ding2024joint}}. Zhang et al. {\cite{zhang2025recurrent}} fine-tuned the visual foundation model FastSAM {\cite{zhao2023fast}} using adapters, while integrating RNNs to model semantic correlations and capture change features. Additionally, some studies have incorporated auxiliary tasks to assist SCD. For instance, Zhou et al. {\cite{zhou2025depthcd}} proposed DepthCD, which automatically models depth change information in remote sensing images and uses it as a cue to mitigate missed detections and false changes caused by shadow occlusion. Tang et al. {\cite{tang2025semantic}} introduced an edge detection task to enhance overall model performance through inter-task interaction. Benefiting from their excellent performance, multi-task methods have become the mainstream approach in current SCD research.

Unlike the aforementioned methods that typically use change information as a post-processing step to enhance spatial consistency, our method focuses on leveraging the guidance of both change and non-change regions at intermediate feature. By incorporating background cues enriched with contextual information, the approach alleviates semantic ambiguity while improving the model’s sensitivity to subtle changes.

\section{LevirSCD}
\label{sec:levirscd}

\begin{figure}
        \centering
        \includegraphics[width=0.42\textwidth]{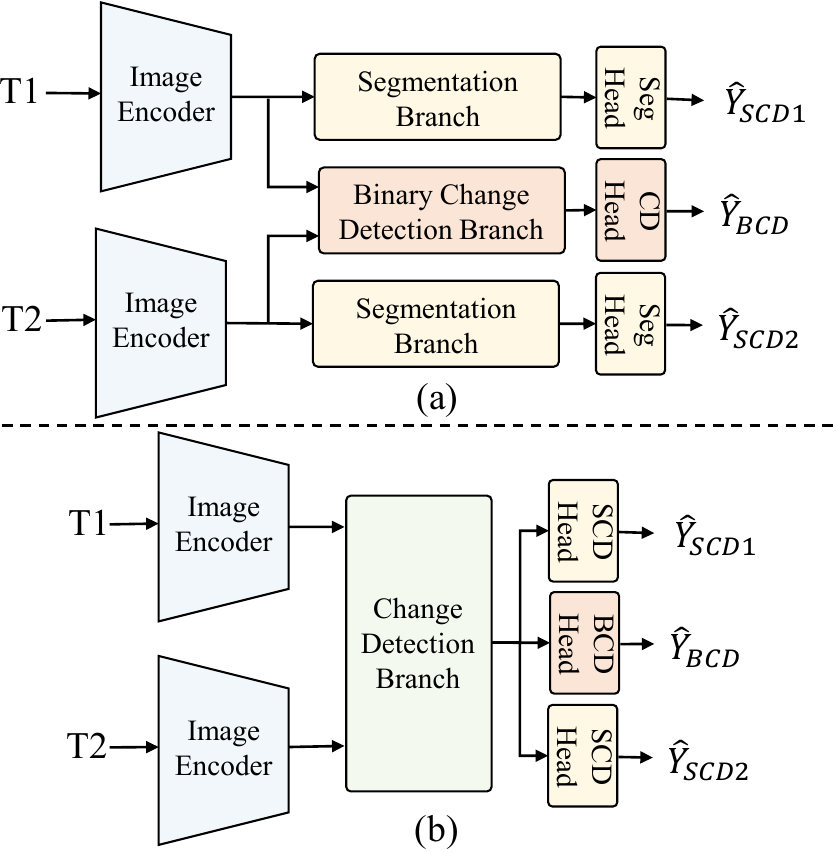}  
        \caption{Illustration of different model construction strategies. (a) The construction strategy used by most existing methods. (b) The approach adopted by our FoBa. T1 and T2 denote the bi-temporal images. The $\hat{Y}_{BCD}$ denotes the prediction results of BCD, while $\hat{Y}_{SCD1}$ and $\hat{Y}_{SCD2}$ are the prediction results of SCD.}
        \label{fig:scd_arch}
\end{figure}

\begin{figure*}
        \centering        \includegraphics[width=0.94\textwidth]{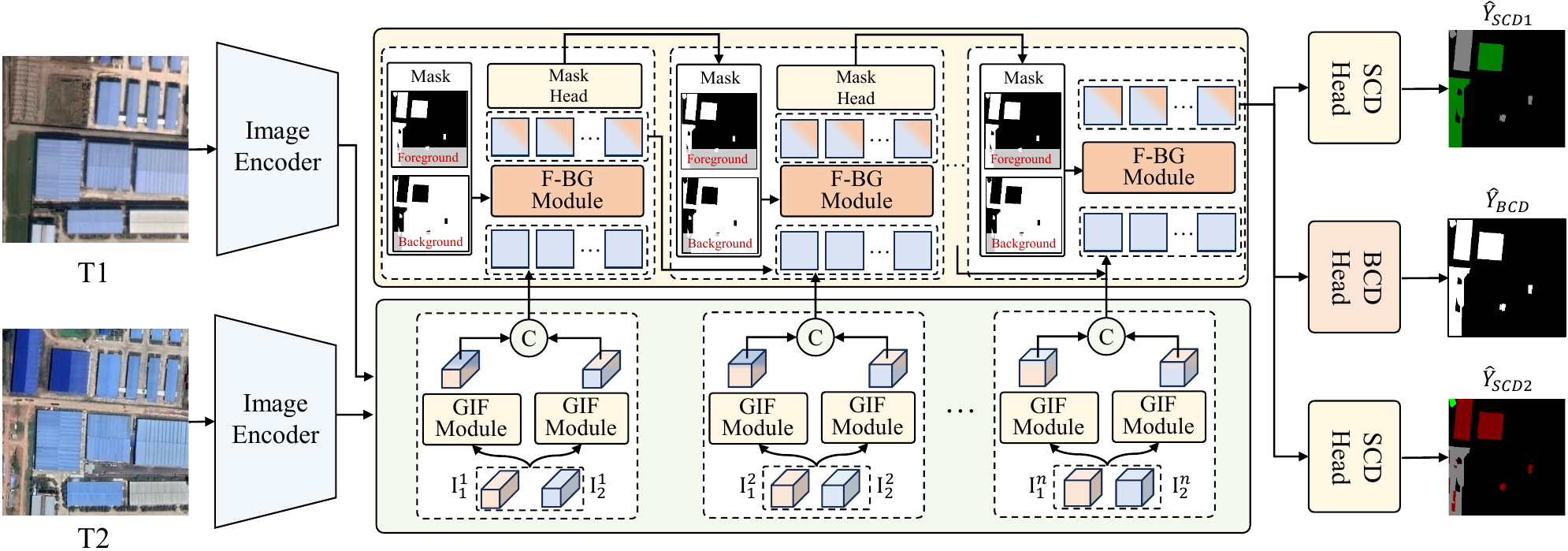}
        \caption{The architecture of the proposed FoBa. The bi-temporal images are fed into the image encoder to extract multi-stage features, which are then passed through multiple GIF modules for bi-temporal feature interaction. The interacted and fused features are subsequently processed by a cascade of F-BG modules to achieve foreground–background co-guidance. Finally, the output of the last F-BG module is fed into different task heads to generate predictions for both BCD and SCD. C denotes the concatenation operation.
        }
        \label{fig:f_bgscd_arch}
\end{figure*}

\begin{figure}
        \centering
        \includegraphics[width=0.4\textwidth]{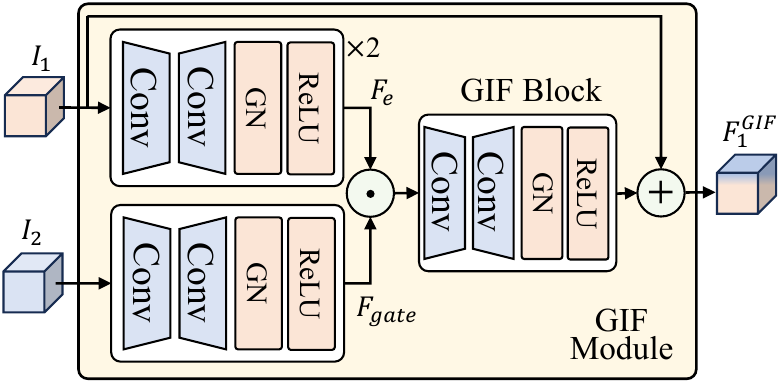} 
        \caption{Illustration of our GIF. The GN denotes group normalization, and $\odot$ represents the hadamard product.
        }
        \label{fig:GIF}
\end{figure}

\subsection{Dataset Construction}
To further advance research on SCD, we constructed an urban scenario SCD dataset, LevirSCD, to support fine-grained change detection tasks. This dataset integrates high-resolution remote sensing imagery from GF-1 and Google Earth, covering the Beijing area with a spatial resolution of 1–2 m/pixel. It spans a temporal range of 9 years (2010–2019) and encompasses a total area of approximately 684 km$^2$. Following mainstream methods such as \cite{LEVIR-CD, WHU-CD}, we partition the original images into non-overlapping patches of size 256 × 256, resulting in 3255 bi-temporal image pairs. During the annotation phase, a team of experts performed pixel-wise manual labeling to ensure the accuracy and consistency of the annotations. Similar to existing SCD datasets, LevirSCD provides the change region masks (binary change mask) as well as semantic change masks from T1 to T2 and from T2 to T1, thereby offering data support for subsequent SCD research.

LevirSCD encompasses a diverse set of change types, comprising 16 typical land cover change categories and 1 unchanged category, for a total of 17 classes. The 16 land cover categories can be further subdivided as follows: 4 natural land types (low vegetation, bare land, woodland, sparse woodland), 2 natural water types (river, water), 4 building types (low building, high building, general building, others), 3 transportation infrastructures (unpaved road, paved road, parking lot) and 3 artificial surfaces (construction land, impervious surfaces, playground). Fig. \ref{fig:levirscd_vis_static} (a) presents some examples from our proposed LevirSCD.

\subsection{Comparison with Other Datasets}

To further highlight the advantages of the proposed LevirSCD dataset, Table \ref{tab:scd_datasets} summarizes the basic characteristics of commonly used open-source SCD datasets, including the number of change categories, number of images, annotation granularity, temporal span, image size, and resolution. Compared with mainstream datasets, LevirSCD contains the largest number of change categories. Specifically, it includes approximately 3 $\times$ as many categories as the SECOND dataset and about 1.5 $\times$ more than WUSU, which previously had the greatest number of categories. In terms of annotation granularity, unlike datasets such as Landsat and JL1, which assign an entire region to a single category (for example, labeling a residential area containing roads and other classes simply as “buildings” without distinguishing the different types), LevirSCD provides separate annotations for the different categories within a region (e.g., individual building instances). Compared with other object-level annotated datasets, such as SECOND and Hi-UCD, LevirSCD provides finer granularity in category classification. For example, roads are subdivided into unpaved road and paved road, while buildings are distinguished as low building, high building. In addition, LevirSCD spans a period of 9 years, enabling the inclusion of more diverse changes. Overall, LevirSCD is a dataset that integrates a diversity of change categories, object-level annotations, and fine-grained semantic class divisions.

\subsection{Dataset Analysis}

To further reveal the statistical characteristics of the LevirSCD dataset, we provides an analysis from three perspectives: sample level, object level, and pixel level. At the sample level, we analyzed the occurrence frequency of each category across the entire dataset. The statistical results are shown in Fig \ref{fig:levirscd_vis_static} (b). It can be observed that low vegetation appears most frequently in the samples, accounting for 20.8\%, followed by low building at 15\% and construction land at 12.3\%. This result reveals the characteristic of rapid residential expansion and indicates that low vegetation and construction land are extensively distributed as transitional forms during the process of urban growth. At the object level, we employed connected component analysis to count the number of instances for each category. Among them, the nature land class contains the largest number of instances, totaling 9437, including 5255 low vegetation, 2425 bare land, 996 woodland, and 761 sparse woodland. The building class follows with 6622 instances, consisting of 4961 low buildings, 817 high buildings, 675 other buildings, and 169 general buildings. The artificial surfaces class includes 4551 instances, comprising 2968 construction land, 1554 impervious surfaces, and 29 playgrounds. The transportation infrastructures class contains 2174 instances, including 1013 unpaved road, 1009 paved road, and 152 parking lots. Finally, the natural water class contains a total of 918 instances, including 564 water bodies and 354 rivers. Overall, each sample contains an average of approximately 7.4 instances. At the pixel level, we conducted a statistics on the number of pixels in each category, and the results are shown in Fig \ref{fig:levirscd_vis_static} (c). It can be observed that low vegetation, construction land, low buildings occur most frequently, ranking among the top three in terms of pixel count, with 1.41 $\times$ $10^7$, 9.75 $\times$ $10^6$, and 5.86 $\times$ $10^6$ pixels, respectively. In contrast, the category with the fewest pixels is playground, at only 5.59 $\times$ $10^4$ pixels, which is approximately 250 times less than that of low vegetation. This obvious class imbalance represents a significant challenge for the dataset.

\section{Foreground-Background co-Guided semantic change detection Method}
\label{sec:method}


\subsection{Overview}
\label{ssec:overview}




Currently, a large number of multi-task based SCD methods (\cite{BiSRNet,li2025enhancing, si2025multi, wang2025refinement}) typically adopt a strategy of decoupling the semantic branch and the binary change detection branch to learn different types of information separately, as illustrated in Fig.\ref{fig:scd_arch} (a). However, these methods exhibit redundancy at both the architectural level, owing to the multi-branch design, and the informational level, as all features are extracted from bi-temporal images. To this end, we propose a more streamlined architecture, as shown in Fig.\ref{fig:scd_arch} (b). Specifically, a single change detection branch aggregates information from bi-temporal images, which is then decoded by multiple task heads. This unified design mitigates both architectural and informational redundancies.

As illustrated in Fig. \ref{fig:f_bgscd_arch}, the proposed FoBa architecture comprises the image encoder, the Gated Interaction Fusion (GIF) module, the Foreground-Background co-Guided (F-BG) module, and task heads. Given the bi-temporal remote sensing images {$\text{T1}$} and {$\text{T2}$}, they are first fed into the image encoder to extract multi-scale features \(\left\{\mathbf{I_1^i}\right\}_{i=1}^4\) and \(\left\{\mathbf{I_2^i}\right\}_{i=1}^4\). The multi-scale features \(\mathbf{I_1^i}\) and \(\mathbf{I_2^i}\) (i = 1,2,3...) are then input into the GIF module to produce guided features \(\mathbf{F_1^{GIF}}\) and \(\mathbf{F_2^{GIF}}\) (for convenience, stage labeling is omitted), which are then concatenated to obtain the fused feature \(\mathbf{F_{out}}\). This process employs gated guidance from the other temporal image to facilitate interaction between the bi-temporal features, thereby enhancing their representations. Subsequently, the \(\mathbf{F_{out}}\) from different stages are sequentially fed into the cascaded F-BG modules, producing \(\mathbf{F_{F-BG}}\), the change-region mask \(\mathbf{M_{c}}\), and the unchanged-region mask \(\mathbf{M_{uc}}\). The outputs \(\mathbf{F_{F-BG}}\), \(\mathbf{M_{c}}\), and \(\mathbf{M_{uc}}\) from the previous stage are then passed as inputs to the next F-BG module. By leveraging guidance from the change and unchanged region masks, this mechanism not only simplifies the learning of change features but also effectively alleviates semantic ambiguities and enhances the detection of subtle changes. Finally, the features output from the last F-BG module are fed into task heads, which decode the binary change mask and the semantic change mask, respectively.

\subsection{Gated Interaction Fusion Module }
\label{ssec:GIF}








Bitemporal feature interaction is a critical component in SCD tasks. By leveraging features from the other temporal as guidance, the model can dynamically integrate information of interest. Based on this idea, we propose a simple yet effective bitemporal feature interaction module, named the GIF module, as illustrated in \ref{fig:GIF}.

Taking the \(\mathbf{I_2}\) guiding \(\mathbf{I_1}\) as an example (for simplicity, stage indices are omitted), given the bi-temporal image features \(\mathbf{I_1}\in \mathbb{R}^{C \times H \times W}\) and \(\mathbf{I_2}\in \mathbb{R}^{C \times H \times W}\), where \(\mathbf{C}\) denotes the channel dimension of the feature maps, \(\mathbf{H}\) and \(\mathbf{W}\) represent the height and width, respectively. The bi-temporal features are processed through the GIF block composed of convolution, GroupNorm, and ReLU. In the GIF block, we first employ a pointwise convolution to project the features into a lower-dimensional space, followed by a depthwise convolution performed within this space, and finally map the results back to the original dimensionality. This bottleneck design effectively reduces the number of model parameters, thereby improving computational efficiency. Unlike \(\mathbf{I_1}\), which obtains the enhanced feature \(\mathbf{F_{e}}\) through two GIF blocks, \(\mathbf{I_2}\) learns the feature \(\mathbf{F_{gate}}\) through a single GIF block.

\begin{equation}
    \label{1}
        \mathbf{F'} = ReLU(GN(Conv_{pdp}(\mathbf{I_1})))))
\end{equation}
\begin{equation}
    \label{2}
        \mathbf{F_{e}} = ReLU(GN(Conv_{pdp}(\mathbf{F'})))))
\end{equation}
\begin{equation}
    \label{3}
        \mathbf{F_{gate}} = ReLU(GN(Conv_{pdp}(\mathbf{I_2})))))
\end{equation}

To aggregate the gated features, \(\mathbf{F_{e}}\) and \(\mathbf{F_{gate}}\) are fused via the hadamard product to obtain the gated feature \(\mathbf{F_{fusion}}\). Subsequently, \(\mathbf{F_{fusion}}\) is further processed by a GIF block and combined with \(\mathbf{I_1}\) through a residual connection, yielding the final output \(\mathbf{F_1^{GIF}}\).

\begin{equation}
    \label{4}
        \mathbf{F_{fusion}} = \mathbf{F_{e}} \odot \mathbf{F_{gate}} 
\end{equation}
\begin{equation}
    \label{5}
        \mathbf{F_1^{GIF}} = ReLU(GN(Conv_{pdp}(\mathbf{F_{fusion}}))))) + \mathbf{I_1}
\end{equation}

Similarly, the result of \(\mathbf{I_1}\) guiding \(\mathbf{I_2}\) denoted as \(\mathbf{F_2^{GIF}}\), can be obtained by swapping the input positions of \(\mathbf{I_1}\) guiding \(\mathbf{I_2}\). Finally, \(\mathbf{F_1^{GIF}}\) and \(\mathbf{F_2^{GIF}}\) are concatenated to form \(\mathbf{F_{out}}\), which serves as the input to the F-BG module.

\begin{figure}
        \centering
        \includegraphics[width=0.5\textwidth]{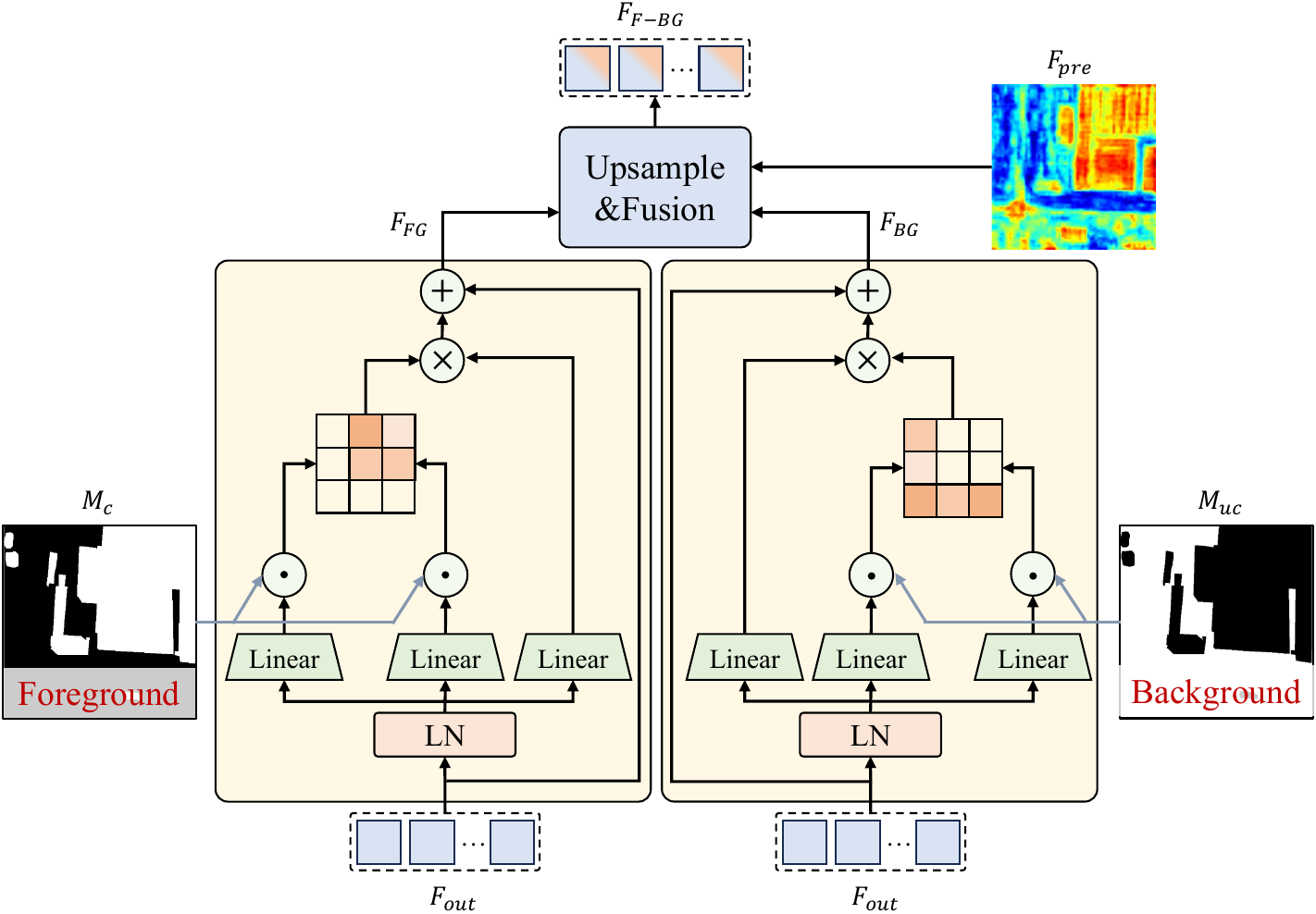} 
        \caption{Illustration of our Transformer-based F-BG module.}
        \label{fig:F_BG_transformer_based}
\end{figure}

\begin{figure}
        \centering
        \includegraphics[width=0.45\textwidth]{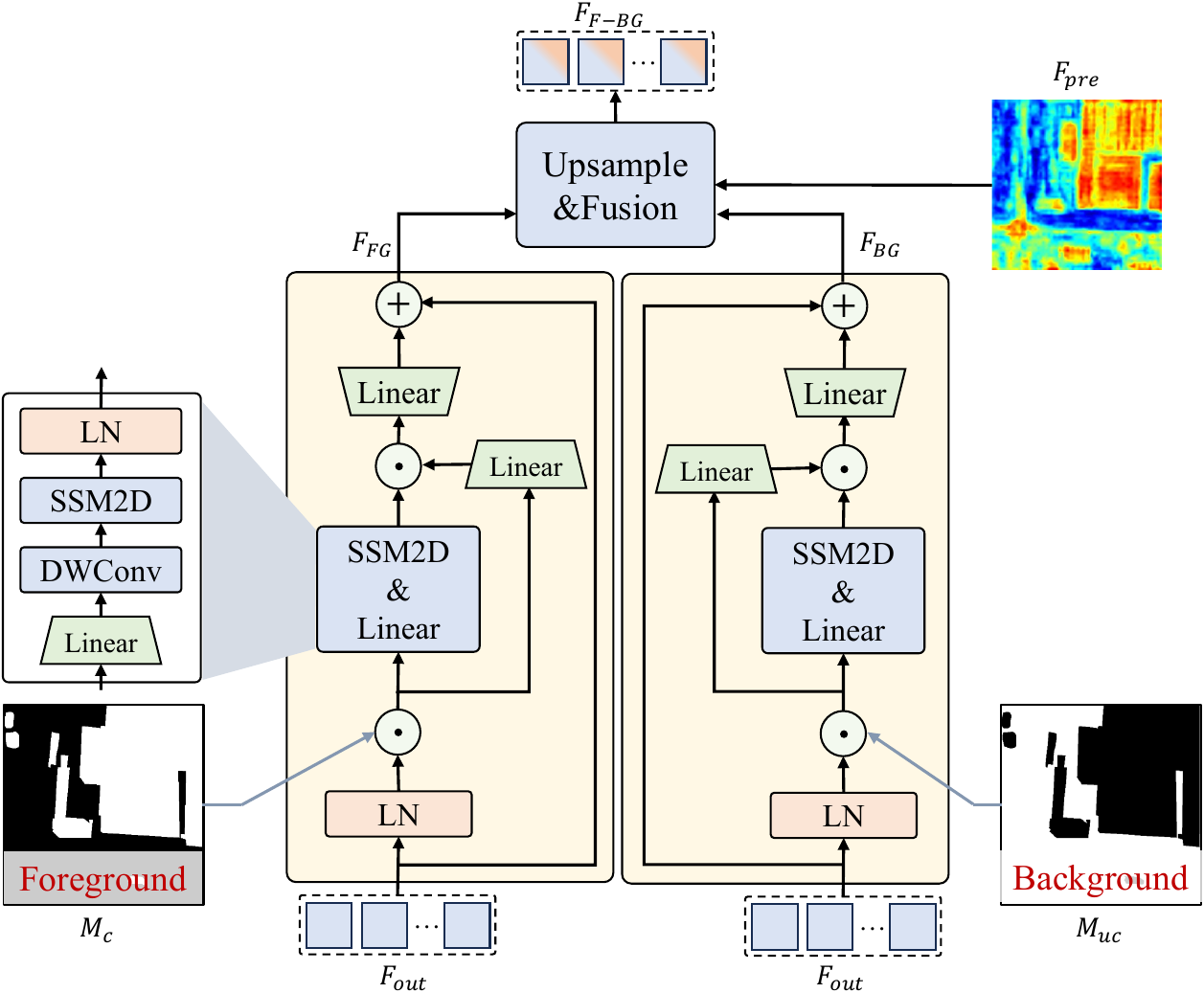} 
        \caption{Illustration of our Mamba-based F-BG module.}
        \label{fig:F_BG_mamba_based}
\end{figure}

\subsection{Foreground-Background co-Guided Module}
\label{ssec:F-BG}









The unchange (background) region contains rich contextual information, which plays a crucial role in enhancing the detection accuracy of SCD tasks, particularly in improving the discrimination of boundary regions. Based on this, we propose two variants, the Transformer-based F-BG module and the Mamba-based F-BG module, which leverage the guidance of foreground and background regions to exploit and utilize change information more effectively. The proposed modules are illustrated in Fig. \ref{fig:F_BG_transformer_based} and Fig. \ref{fig:F_BG_mamba_based}, respectively.

\textbf{Transformer-based F-BG:} Taking the second stage as an example, where the fused feature \(\mathbf{F_{out}}\) (stage indices omitted for clarity) is simultaneously fed into the F-G and B-G blocks. Within the F-G block, \(\mathbf{F_{out}}\) is first normalized via Layer Normalization, after which convolutional projections are employed to derive the query \(\mathbf{Q}\), key \(\mathbf{K}\), and value \(\mathbf{V}\) vectors. Notably, in contrast to the original self-attention mechanism that relies on fully connected layer, we adopt the convolution-based projection strategy introduced in \cite{zamir2022convtransformer}. 

\begin{equation}
    \label{6}
[\mathbf{Q}, \mathbf{K}, \mathbf{V}] = Conv_1(Conv_3(\mathbf{F_{out}} \otimes \mathbf{I}_3)) 
\end{equation}
where $Conv_{1}$ and $Conv_{3}$ denote convolution operations with kernel sizes of 1 and 3, respectively, while \(\mathbf{I}_3\) indicates replicating \(\mathbf{F_{out}}\) into three identical copies. Next, the change-region mask \(\mathbf{M_c}\) is applied to the \(\mathbf{Q}\) and \(\mathbf{K}\) via element-wise multiplication to obtain the masked query \(\mathbf{Q_M}\) and masked key \(\mathbf{K_M}\). These are then used to construct the attention matrix, which is applied to the \(\mathbf{V}\) to produce the foreground attention feature \(\mathbf{F_{FA}}\). This mechanism explicitly decouples foreground information while suppressing background interference, thereby simplifying the learning process.

\begin{equation}
    \label{7}
(\mathbf{Q_M}, \mathbf{K_M}) = (\mathbf{Q} \odot \mathbf{M_c}, \;\mathbf{K} \odot \mathbf{M_c})
\end{equation}
\begin{equation}
    \label{8}
        \mathbf{F_{FA}} = Softmax \left(\frac{\mathbf{Q_M} \mathbf{K_M}^{T}}{\alpha}\right) \mathbf{V}
\end{equation}

 Finally, \(\mathbf{F_{FA}}\) is processed through a residual connection and a feed-forward network, as in standard self-attention, to yield the foreground-guided feature \(\mathbf{F_{FG}}\).

Following the same procedure as \(\mathbf{F_{FG}}\), the background-guided feature \(\mathbf{F_{BG}}\) is obtained by replacing \(\mathbf{M_c}\) with \(\mathbf{M_{uc}}\), enriching contextual information. \(\mathbf{F_{FG}}\), \(\mathbf{F_{BG}}\) and the previous F-BG module output \(\mathbf{F_{pre}}\) are then fused via element-wise addition and convolution to produce the foreground–background guided feature \(\mathbf{F_{F-BG}}\).

Notably, at each stage, \(\mathbf{M_c}\) is obtained by applying convolution followed by a sigmoid activation to \(\mathbf{F_{F-BG}}\), while \(\mathbf{M_{uc}}\) is obtained as \(\mathbf{1} - \mathbf{M_c}\).  At the initial stage, \(\mathbf{M_c}\) is directly generated from \(\mathbf{F_{out}}\) via convolution and sigmoid.



\textbf{Mamba-based F-BG:} Furthermore, based on the same principle, we propose a Mamba-based variant. The fused feature \(\mathbf{F_{out}}\) is first normalized via Layer Normalization and combined with \(\mathbf{M_c}\) through element-wise multiplication to produce the masked feature \(\mathbf{F_{Masked}}\). The \(\mathbf{F_{Masked}}\) is then split into two branches: one passes sequentially through Linear, DWConv, SS2D, and Layer Normalization, while the other is processed by Linear and fused back into the first branch via element-wise multiplication.

\begin{equation}
    \label{9}
        \mathbf{F_{Masked}} = LN(\mathbf{F_{out}}) \odot \mathbf{M_c}
\end{equation}
\begin{equation}
    \label{10}
        \mathbf{F_{b1}} = LN(SS2D(DWConv(Linear(\mathbf{F_{Masked}}))))
\end{equation}
\begin{equation}
    \label{11}
        \mathbf{F_{b2}} = Linear(\mathbf{F_{Masked}})
\end{equation}
\begin{equation}
    \label{12}
        \mathbf{F_{fusion}} = \mathbf{F_{b1}} \odot \mathbf{F_{b2}}
\end{equation}

Following the Transformer-based approach, the \(\mathbf{F_{FG}}\) is produced via a residual connection and feed-forward network, while the \(\mathbf{F_{BG}}\) is obtained by replacing \(\mathbf{M_c}\) with \(\mathbf{M_{uc}}\). \(\mathbf{F_{FG}}\), \(\mathbf{F_{BG}}\), and the output of the previous F-BG module are then fused using the same strategy to yield the \(\mathbf{F_{F-BG}}\).

\subsection{Loss Functions}
\label{ssec:loss}


To effectively train the proposed method, the overall loss function \( L_{total}\) is composed of 5 components: the binary change detection classification loss \( L_{bcd}\), the semantic change detection classification loss \( L_{scd}\), the sample imbalance loss \( L_{sample}\), the foreground loss \( L_{f}\) and the non-change region semantic consistency loss \( L_{cons}\).

Specifically, \( L_{bcd}\) and \( L_{scd}\) are optimized using the cross-entropy loss. In addition, the Lovasz-softmax loss \cite{berman2018lovasz} is added to mitigate class imbalance between change and non-change pixels. The  \( L_{f}\) at each stage is constrained via binary cross-entropy due to the F-BG strategy. Furthermore, considering the semantic consistency characteristic of the semantic change detection task, we introduce the \( L_{cons}\), which imposes a mean-squared error constraint on non-change regions, ensuring consistent semantic predictions and thereby enhancing overall semantic modeling. Finally, the total loss is formulated as: 
\begin{equation}
    \label{13}
         L_{total} = \lambda_1{L_{bcd}}+\lambda_2({L_{scd}}+L_{cons})+\lambda_3{L_{sample}}+\lambda_4{L_{f}}
\end{equation}
where \( \lambda_i\) are hyperparameters controlling the contributions of each loss term. Their effects are further analyzed in the ablation study.

\section{Experimental Results and Analysis} 
\label{sec:experiment}

\subsection{Data description}
\label{ssec:data}





Extensive experiments are conducted on three representative datasets (SECOND, JL1, and the proposed LevirSCD) to validate the effectiveness of the proposed FoBa method comprehensively.

SECOND\cite{yang2021asymmetric}: The SECOND dataset comprises 4662 image pairs collected from multiple sensors and platforms, covering representative urban regions such as Hangzhou, Shanghai, and Chengdu. Each image has a size of 512 $\times$ 512 pixels, with a spatial resolution ranging from 0.5-3 m/pixel. For all pairs, semantic change annotations are provided in both directions, i.e., from $T_1$ to $T_2$ and from $T_2$ to $T_1$. The dataset encompasses six change categories: non-vegetated ground surfaces, trees, low vegetation, water, buildings, and playgrounds, and includes 30 types of changes. Following the official split, 2968 pairs are allocated for training and 1694 pairs for testing.

LevirSCD: The LevirSCD dataset consists of 3255 images of size 256 $\times$ 256, collected from multiple platforms over the Beijing region, with a spatial resolution ranging from 1-2 m/pixel and spanning a temporal range of 9 years. It covers 16 change categories (e.g., low vegetation, bare land) and encompasses a total of 210 different change types. Following \cite{yang2021asymmetric}, 2580 images are allocated for training and 645 images for testing. A more detailed description and analysis of the dataset can be found in Sec. \ref{sec:levirscd}.

JL1: The JL1 dataset is a competition benchmark designed for farmland-type CD, comprising 6000 images acquired from the Jilin-1 remote sensing satellite. Each image has a spatial resolution of 0.75 m/pixel with a size of 256 $\times$ 256. Unlike LevirSCD and SECOND, which provide bidirectional "from-to" annotations, JL1 adopts a single-label annotation covering 8 change types: cropland-building, cropland-forest/grassland, cropland-road, cropland-other, building-cropland, forest/grassland-cropland, road-cropland, and other-cropland. Following the preprocessing strategy in \cite{cdsc}, we convert these labels into the "from-to" format before training. Finally, 4050 images are used for training and 1950 images for testing.

\subsection{Experimental setup}
\label{ssec:setup}

\subsubsection{Architecture details}
\label{ssec:architecture}

We adopt the base version of the Mamba-based model \cite{liu2024vmamba} as the image encoder, where the feature maps across its four stages are downsampled to scales of 1/4, 1/8, 1/16, and 1/32 of the original image size. The channel dimensions of the GIF modules at different stages are set to 128, 256, 512, and 1024, respectively, while those of all F-BG modules are uniformly set to 128. To generate foreground-guided features, we employ a mask head implemented with a convolution layer, which takes 128 input channels and produces a single-channel output with a kernel size of 3 and a padding of 1. In addition, both the BCD head and the SCD head are implemented with a single convolutional layer of kernel size 1, where the output channels are set to 2 and the number of classes defined in the dataset, respectively.

\subsubsection{Training details}
\label{ssec:training}


The proposed FoBa is implemented on the PyTorch framework and trained on a single NVIDIA RTX 4090 GPU. We adopt the AdamW optimizer \cite{adamw} with a learning rate of 1$e$-4 and a weight decay of 5$e$-4. The batch size is set to 2 for the SECOND dataset and 6 for both LevirSCD and JL1 datasets. Regarding training iterations, we use 480k steps for SECOND and 600k steps for LevirSCD and JL1. The detailed configuration of the loss functions is provided in Sec. \ref{ssec:loss}.

\subsubsection{Evaluation metrics}
\label{ssec:evaluation}

In this study, we employed two categories of evaluation metrics to assess the overall accuracy of the SCD task. These include one BCD metric, mean intersection over union (mIoU), and three SCD metrics: overall accuracy (OA), Separated Kappa (SeK) coefficient, and $F_{scd}$. Let the confusion matrix be denoted as $Q = \{ q_{i,j} \}$ where $q_{i,j}$ represents the number of pixels classified into class $i$ while their ground-truth label is $j$ \((i,j \in \{1,2,3,...N\})\), with the unchanged class assigned to 0. The computation process of the mIoU is described as follows:

\begin{equation}
    \label{13}
        IoU_{nc} = q_{00} / (\sum_{i=0}^{N}q_{i0}+\sum_{j=0}^{N}q_{0j} - q_{00}) 
\end{equation}
\begin{equation}
    \label{14}
        IoU_{c} = \sum_{i=1}^{N}\sum_{j=1}^{N}q_{ij} / (\sum_{i=0}^{N}\sum_{j=0}^{N}q_{ij} - q_{00}) 
\end{equation}
\begin{equation}
    \label{15}
        mIoU = (IoU_{nc}+IoU_{c})/2
\end{equation}
And, the computation of OA is given as follows:
\begin{equation}
    \label{13}
        OA = \sum_{i=0}^{N} q_{ii} / \sum_{i=0}^{N}\sum_{j=0}^{N} q_{ij}
\end{equation}
The computation of SeK is based on the confusion matrix $\hat{Q} = \{\hat{q}_{ij}\}$ , where $\hat{q}_{ij}$ = $q_{i,j}$ except that $\hat{q}_{00}$ = 0. This adjustment eliminates the influence of the dominant true positive unchange pixels. The specific computation procedure is as follows:
\begin{equation}
    \label{14}
        \rho = \sum_{i=1}^{N} \hat{q}_{ii} / \sum_{i=0}^{N}\sum_{j=0}^{N} \hat{q}_{ij}
\end{equation}
\begin{equation}
    \label{15}
        \eta = (\sum_{i=1}^{N}(\sum_{j=0}^{N} \hat{q}_{ij} \ast \sum_{j=0}^{N} \hat{q}_{ji})) / (\sum_{i=0}^{N}\sum_{j=0}^{N}\hat{q}_{ij})^{2}
\end{equation}
\begin{equation}
    \label{16}
        SeK = e^{IoU_{c}-1} \cdot (\rho-\eta)/(1-\eta)
\end{equation}
The computation of $F_{scd}$ is as follows:
\begin{equation}
    \label{17}
        P_{scd} = \sum_{i=1}^{N}{q}_{ii} / \sum_{i=1}^{N}\sum_{j=0}^{N} {q}_{ij}
\end{equation}
\begin{equation}
    \label{18}
        R_{scd} = \sum_{i=1}^{N}{q}_{ii} / \sum_{i=0}^{N}\sum_{j=1}^{N} {q}_{ij}
\end{equation}
\begin{equation}
    \label{19}
        F_{scd} = \frac{2 \ast P_{scd} \ast R_{scd}}{P_{scd}+R_{scd}} 
\end{equation}
where $P_{scd}$ and $R_{scd}$ are variants of precision and recall computed exclusively over the change regions. Accordingly, $F_{scd}$ reflects the detection accuracy specifically within these change regions.

Since OA is easily biased by the large number of unchanged pixels in SCD tasks, it cannot reliably reflect detection performance. Accordingly, SeK, which excludes the influence of no-change pixels, and $F_{scd}$ , which focuses on the change regions, serve as key metrics for evaluating the semantic accuracy of SCD tasks. Meanwhile, mIoU, reflecting the segmentation accuracy of both change and unchanged regions, is also an important metric of interest.

\begin{table*}
    \centering
    \caption{Comparison results on the three SCD test sets. The top three results are highlighted in \colorbox{myred}{red}, \colorbox{myorg}{orange}, \colorbox{myyellow}{yellow}. All results are described in percentage (\%).}
    \resizebox{1\textwidth}{!}{
    \begin{tabular}{c|c|c|c|c}
    \toprule
    \multicolumn{1}{c|}{} &
    \multirow{2}{*}{\textbf{Backbone}}  &
    \multicolumn{1}{c|}{\textbf{SECOND}}  &
    \multicolumn{1}{c|}{\textbf{LevirSCD}}  &
    \multicolumn{1}{c}{\textbf{JL1}}  \\
    & & F$_{\text{scd}}$ / mIoU / Sek$_{\text{37}}$ / OA & F$_{\text{scd}}$ / mIoU / Sek$_{\text{257}}$ / OA &  F$_{\text{scd}}$ / mIoU / Sek$_{\text{26}}$ / OA \\
    \midrule
    HRSCD-str4$_{19}$ \cite{hrscd4}
    & CNN
    & \colorbox{mywhite}{49.03} / \colorbox{mywhite}{71.08} / \colorbox{mywhite}{17.71} / \colorbox{mywhite}{84.42}   
    & \colorbox{mywhite}{20.36} / \colorbox{mywhite}{73.87} / \colorbox{mywhite}{ 6.21} / \colorbox{mywhite}{85.47} 
    & \colorbox{mywhite}{59.43} / \colorbox{mywhite}{76.73} / \colorbox{mywhite}{29.80} / \colorbox{mywhite}{83.23} \\
    SSCDl$_{22}$ \cite{BiSRNet} 
    & CNN
    & \colorbox{mywhite}{54.07} / \colorbox{mywhite}{73.22} / \colorbox{mywhite}{21.86} / \colorbox{mywhite}{85.83}  
    & \colorbox{mywhite}{34.91} / \colorbox{mywhite}{74.07} / \colorbox{mywhite}{14.21} / \colorbox{mywhite}{88.00} 
    & \colorbox{mywhite}{84.55} / \colorbox{mywhite}{86.53} / \colorbox{mywhite}{59.32} / \colorbox{mywhite}{92.69}\\
    BiSRNet$_{22}$ \cite{BiSRNet} 
    & CNN
    & \colorbox{mywhite}{53.98} / \colorbox{mywhite}{73.10} / \colorbox{mywhite}{21.66} / \colorbox{mywhite}{85.88}  
    & \colorbox{mywhite}{36.95} / \colorbox{mywhite}{74.67} / \colorbox{mywhite}{15.71} / \colorbox{mywhite}{88.19} 
    & \colorbox{mywhite}{84.93} / \colorbox{mywhite}{86.62} / \colorbox{mywhite}{59.78} / \colorbox{mywhite}{92.75} \\
    TED$_{24}$ \cite{TED} 
    & CNN
    & \colorbox{mywhite}{54.03} / \colorbox{mywhite}{73.27} / \colorbox{mywhite}{22.05} / \colorbox{mywhite}{85.56}  
    & \colorbox{mywhite}{35.32} / \colorbox{mywhite}{74.64} / \colorbox{mywhite}{14.81} / \colorbox{mywhite}{88.16} 
    & \colorbox{mywhite}{83.80} /
    \colorbox{mywhite}{86.41} / \colorbox{mywhite}{58.53} / \colorbox{mywhite}{92.37} \\
    SCanNet$_{24}$ \cite{TED} 
    & CNN
    & \colorbox{mywhite}{55.29} / \colorbox{mywhite}{73.43} / \colorbox{mywhite}{22.27} / 
    \colorbox{myred}{\textbf{86.26}} 
    & \colorbox{mywhite}{37.39} / \colorbox{mywhite}{74.95} / \colorbox{mywhite}{16.11} / \colorbox{mywhite}{88.49} 
    & \colorbox{mywhite}{87.42} / \colorbox{mywhite}{87.60} / \colorbox{mywhite}{63.42} / \colorbox{mywhite}{93.62} \\
    DEFO$_{24}$ \cite{DEFO}  
    & CNN
    & \colorbox{mywhite}{54.42} / \colorbox{myyellow}{\textbf{73.70}} / \colorbox{mywhite}{22.71} / \colorbox{mywhite}{85.65} 
    & \colorbox{mywhite}{34.79} / \colorbox{mywhite}{74.64} / \colorbox{mywhite}{14.49} / \colorbox{mywhite}{88.03} 
    & \colorbox{mywhite}{85.26} / \colorbox{mywhite}{86.93} / \colorbox{mywhite}{60.44} / \colorbox{mywhite}{92.98}\\
    CdSCNet$_{24}$ \cite{cdsc}  
    & Transformer
    & \colorbox{mywhite}{55.22} / \colorbox{mywhite}{73.26} / \colorbox{mywhite}{22.60} / \colorbox{mywhite}{86.19} 
    & \colorbox{mywhite}{38.21} / \colorbox{mywhite}{75.29} / \colorbox{myyellow}{\textbf{16.72}} / \colorbox{mywhite}{88.56} 
    & \colorbox{mywhite}{83.74} / \colorbox{mywhite}{83.45} / \colorbox{mywhite}{54.32} / \colorbox{mywhite}{91.73}\\
    ChangeMamba$_{24}$ \cite{chen2024changemamba} 
    & Mamba
    & \colorbox{mywhite}{55.29} / \colorbox{mywhite}{73.47} / \colorbox{mywhite}{22.86} / \colorbox{myyellow}{\textbf{86.21}}
    & \colorbox{myyellow}{\textbf{38.62}} / \colorbox{myyellow}{\textbf{76.03}} / \colorbox{mywhite}{17.51} / 
    \colorbox{myorg}{\textbf{89.45}}
    & \colorbox{myyellow}{\textbf{88.41}} / \colorbox{myyellow}{\textbf{87.75}} / \colorbox{myyellow}{\textbf{64.57}} / \colorbox{myyellow}{\textbf{93.91}}\\
    LSAFNet$_{25}$ \cite{zhou2024late}  
    & CNN
    & \colorbox{myyellow}{\textbf{56.08}} / \colorbox{mywhite}{73.65} / \colorbox{myyellow}{\textbf{23.13}} / \colorbox{myorg}{\textbf{86.23}} 
    & \colorbox{mywhite}{35.82} / \colorbox{mywhite}{75.24} / \colorbox{mywhite}{15.55} / \colorbox{mywhite}{88.14} 
    & \colorbox{mywhite}{86.40} / \colorbox{mywhite}{85.81} / \colorbox{mywhite}{59.87} / \colorbox{mywhite}{92.97}\\
    \midrule
    Ours(Transformer-based) 
    & Mamba
    & \colorbox{myorg}{\textbf{57.36}} / \colorbox{myorg}{\textbf{74.36}} / \colorbox{myorg}{\textbf{24.57}} / \colorbox{mywhite}{85.99}  
    & \colorbox{myorg}{\textbf{40.82}} / \colorbox{myred}{\textbf{76.99}} / \colorbox{myorg}{\textbf{19.33}} / \colorbox{myyellow}{\textbf{89.28}}
    & \colorbox{myred}{\textbf{89.90}} / \colorbox{myorg}{\textbf{89.19}} / \colorbox{myorg}{\textbf{68.14}} / \colorbox{myorg}{\textbf{94.59}}\\
    Ours(Mamba-based) 
    & Mamba
    & \colorbox{myred}{\textbf{57.57}} / \colorbox{myred}{\textbf{74.50}} / \colorbox{myred}{\textbf{24.61}} / \colorbox{mywhite}{86.14} 
    & \colorbox{myred}{\textbf{41.27}} / \colorbox{myorg}{\textbf{76.82}} /  \colorbox{myred}{\textbf{19.53}} / \colorbox{myred}{\textbf{89.76}}
    & \colorbox{myorg}{\textbf{89.75}} / \colorbox{myred}{\textbf{89.30}} /  \colorbox{myred}{\textbf{68.18}} / \colorbox{myred}{\textbf{94.62}}\\
   \bottomrule
    \end{tabular}
    }
    \label{tab:comparison_sotas}
\end{table*}

\subsection{Performance comparison}
\label{ssec:performance}


To evaluate the effectiveness of the proposed FoBa in the SCD task, several state-of-the-art models are chosen as the competitors, including the CNN-backbone-based methods (HRSCD-str4\cite{hrscd4}, SSCDl\cite{BiSRNet}, BiSRNet\cite{BiSRNet}, TED\cite{TED}, SCanNet\cite{TED}, DEFO\cite{DEFO}, LSAFNet\cite{zhou2024late}), the Transformer-backbone-based method (CdSCNet\cite{cdsc}), and the Mamba-backbone-based method (ChangeMamba\cite{chen2024changemamba}). All competing methods are trained using their official open-source PyTorch implementations.

\subsubsection{Quantitative results}
\label{ssec:quantitative}



In terms of quantitative results, Table \ref{tab:comparison_sotas} summarizes the overall performance of all methods on the SECOND, LevirSCD, and JL1 test sets, where the \colorbox{myred}{red} labels represent the optimal results, and the \colorbox{myorg}{orange} and \colorbox{myyellow}{yellow} correspond to the suboptimal and third-best results respectively.

Compared with the current state-of-the-art approaches, both our Transformer-based and Mamba-based FoBa variants achieve competitive results. Specifically, on the LevirSCD and JL1 datasets, compared with DEFO using CNN-backbone and the outstanding LSAFNet, our method has achieved outperformance in all evaluation metrics. In particular, the SeK and $F_{scd}$ surpass them by 5.04\%/7.74\%, 6.48\%/4.49\%, respectively. On the SECOND dataset, although our approach yields a slightly lower OA than LSAFNet, it achieves notable improvements of 1.48\% and 1.49\% on the more critical SCD metrics SeK and $F_{scd}$. Moreover, in terms of mIoU, which better reflects the accuracy of detected change regions, our method also surpasses LSAFNet by 0.85\%. Compared with the Transformer-backbone-based CdSCNet, our proposed approach achieves substantial improvements across all three datasets, with SeK gains of 2.01\%/2.81\%/13.86\%. Furthermore, relative to ChangeMamba, which also adopts Mamba as its backbone, our method demonstrates strong competitiveness. For instance, on the SECOND and LevirSCD datasets, the SeK and $F_{scd}$ metrics increased by 1.75\%/2.02\% and 2.28\%/2.65\%, respectively.

In summary, the above quantitative analysis indicates the critical role of jointly leveraging foreground and background information in SCD tasks. This strategy not only enables more precise semantic discrimination but also provides a more accurate estimation of change regions.

\subsubsection{Qualitative results}
\label{ssec:qualitative}

\begin{figure*}
    \centering
    \includegraphics[width=0.98\textwidth]{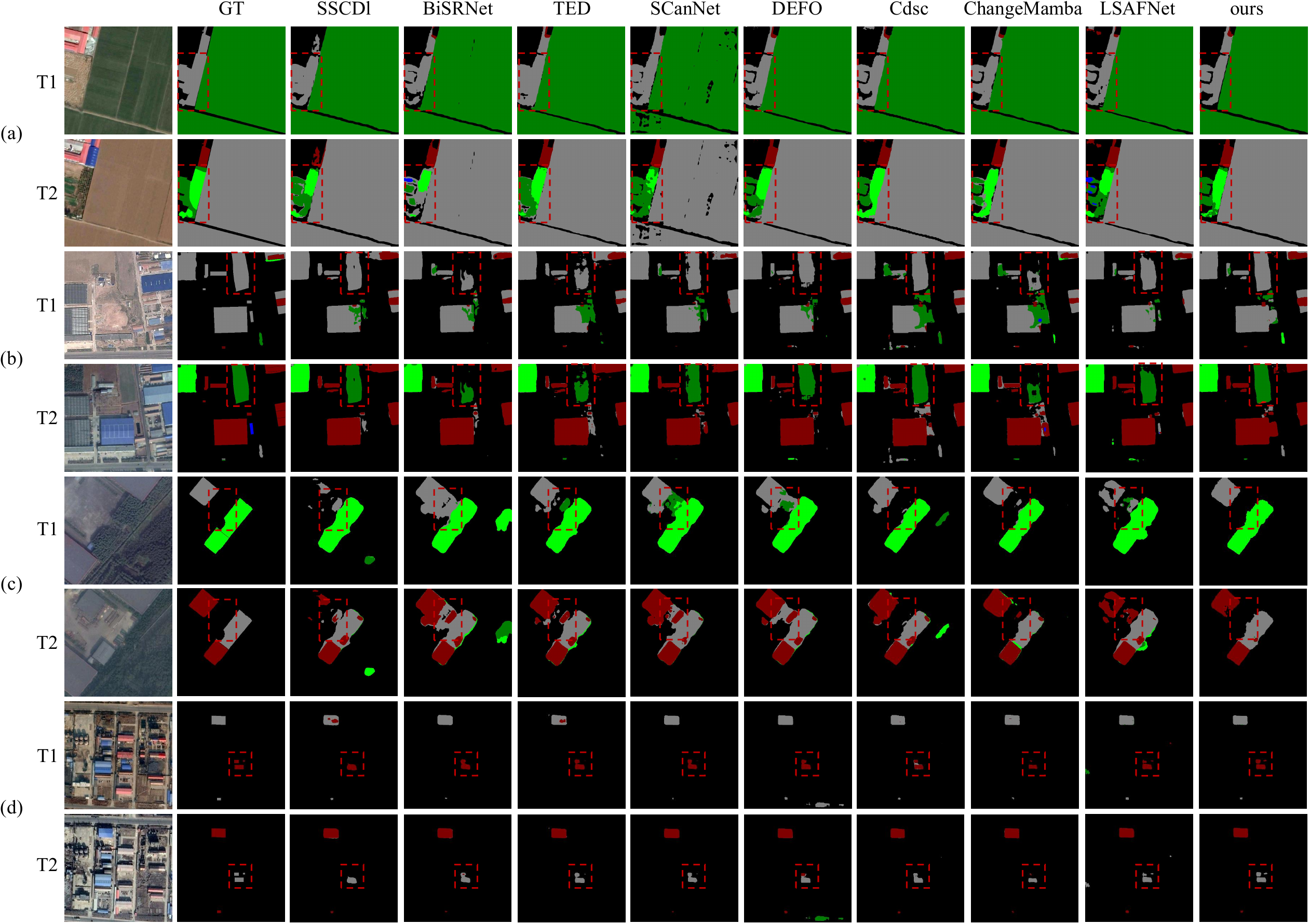}
    \caption{Visualization results of different methods on the SECOND test set. (a)-(d) are representative samples. The black denotes unchanged areas, while the red bounding boxes highlight regions where our method achieves superior results.
    }
    \label{fig:SECOND}
\end{figure*}

\begin{figure*}
    \centering
    \includegraphics[width=0.98\textwidth]{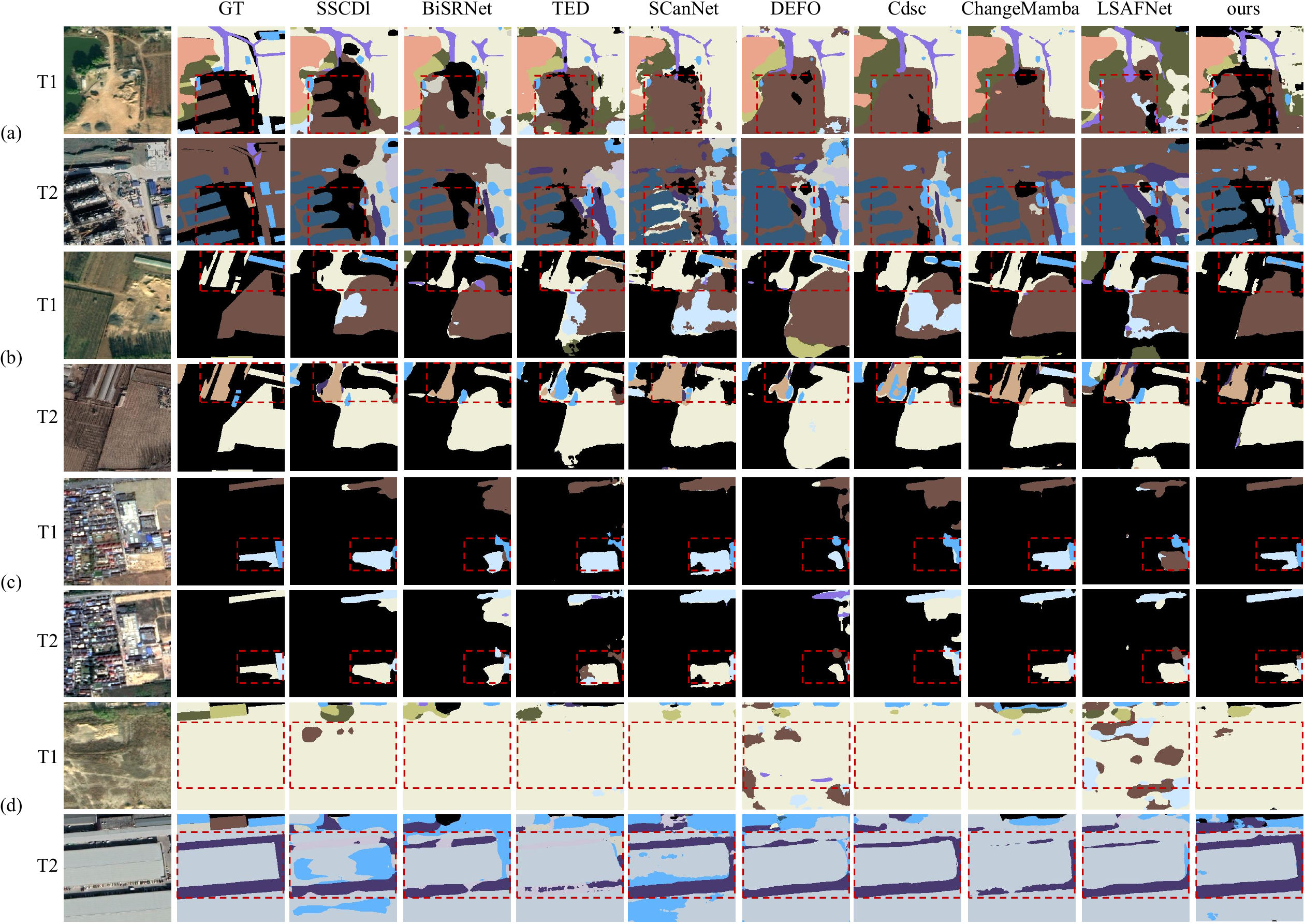}
    \caption{Visualization results of different methods on the LevirSCD test set. (a)-(d) are representative samples. The black denotes unchanged areas, while the red bounding boxes highlight regions where our method achieves superior results.}
    \label{fig:levirscd}
\end{figure*}

\begin{figure*}
    \centering
    \includegraphics[width=0.98\textwidth]{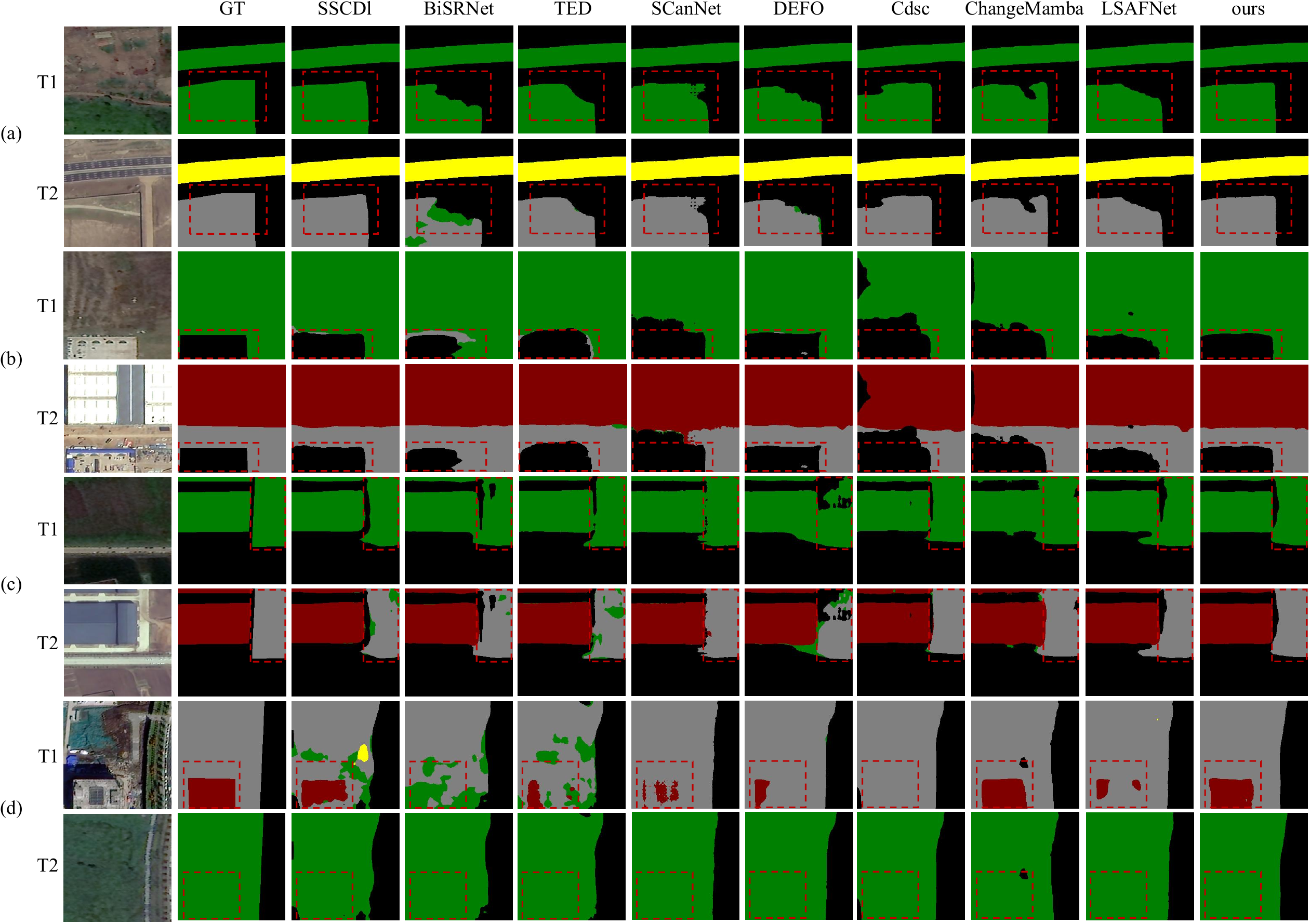}
    \caption{Visualization results of different methods on the JL1 test set. (a)-(d) are representative samples. The black denotes unchanged areas, while the red bounding boxes highlight regions where our method achieves superior results.}
    \label{fig:jl1}
\end{figure*}

To further validate the effectiveness of the proposed approach, we conduct qualitative analyses on SECOND, LevirSCD, and JL1, as illustrated in Fig. \ref{fig:SECOND}–\ref{fig:jl1}. In these visual comparisons, the red boxes highlight regions with pronounced differences.

Visualization on SECOND (Fig. \ref{fig:SECOND}): We select several representative samples, as shown in Fig. \ref{fig:SECOND} (a)-(d). In Fig. \ref{fig:SECOND} (a), although most methods achieve reasonably accurate recognition over large change regions (e.g., low vegetation and ground area), our approach produces more precise semantic detection in the local region marked by the red boxes. Compared with both the high-performing LSAFNet and the Mamba-backbone-based ChangeMamba, our method demonstrates superior accuracy, qualitatively validating the effectiveness of foreground-background co-guidance in mitigating semantic ambiguities. The same advantage is observed in Fig. \ref{fig:SECOND} (b), where, compared to existing methods, our results exhibit more complete object shapes and clearer boundaries. In Fig. \ref{fig:SECOND} (c), compared with almost all methods that produce false detection in the red-boxed regions, our approach effectively integrates context-rich background information, substantially mitigating such misdetections. Moreover, in the case of subtle changes illustrated in Fig. \ref{fig:SECOND} (d), our method outperforms others. Unlike SSCDl and ChangeMamba, which exhibit evident omission errors, or the remaining methods, which suffer from severe adhesion between objects, our approach not only accurately detects the changes but also maintains almost no adhesion between objects. Notably, our method is also capable of accurately detecting the extremely subtle change in the upper right corner marked by the red box. This result provides an intuitive validation of the effectiveness of the foreground–background co-guidance strategy in enhancing the model’s sensitivity to subtle changes. In summary, the qualitative results are consistent with the quantitative analysis in Table \ref{tab:comparison_sotas}, demonstrating that the proposed method achieves highly competitive performance on the SECOND dataset.

Visualization on LevirSCD (Fig. \ref{fig:levirscd}): Similarly, we selected several representative samples from the LevirSCD dataset for illustration, as shown in Fig. \ref{fig:levirscd} (a)-(d). In the complex change scenario shown in Fig. \ref{fig:levirscd} (a), compared to methods such as ChangeMamba and LSAFNet, which exhibit substantial false detection, our method delineates the change regions more accurately. Moreover, compared to SSCDl, which also achieves reasonable segmentation of change areas, our approach demonstrates notably reduced boundary adhesion. In the large change region illustrated in Fig. \ref{fig:levirscd} (b) and (d), it can be observed that, compared to other methods, our approach produces more regular shapes and achieves more precise category delineation. This observation further validates the effectiveness of the foreground–background co-guided strategy in SCD tasks. Moreover, in the low-resolution scenario (2m/pixel) shown in Fig. \ref{fig:levirscd} (c), our method also demonstrates strong performance, producing change shapes and categories that closely align with the ground truth. This result highlights the adaptability of the proposed approach across varying resolution scenarios.

Visualization on JL1 (Fig. \ref{fig:jl1}): In the JL1 dataset, we also selected some representative samples, as shown in Fig. \ref{fig:jl1} (a)-(d). As shown in Fig. \ref{fig:jl1} (a) and (b), while most existing methods produce irregular shapes within the red-boxed regions, our approach achieves detection results that are nearly consistent with the ground truth. In Fig. \ref{fig:jl1} (c), compared to methods such as SCCDl, TED, and DEFO, which exhibit substantial class misclassifications within the change regions, our approach, leveraging the foreground–background co-guided strategy, shows almost no class errors within the red-boxed area and significantly reduces the adhesion phenomenon between different targets. The detection results in the buildings region shown in Fig. \ref{fig:jl1} (d) are consistent with the observations above, further confirming the effectiveness of the foreground–background co-guided strategy in SCD tasks.

\begin{table}
    \centering
    \caption{Comparison results on model efficiency. We report the number of parameters (Params.), and floating-point operations per SECOND (FLOPs). The size of the input image to the model is $512\times 512 \times 3$ to calculate the FLOPs.}
    \begin{tabular}{c|cc}
    \toprule
    Model & Params. (M) & FLOPs (G) \\
    \midrule
        HRSCD-str4$_{19}$  & 13.71 & 43.97 \\
        SSCDl$_{22}$  & 23.31 & 189.76 \\
        BiSRNet$_{22}$ & 23.38 & 190.30 \\
        TED$_{24}$ & 24.19 & 204.29 \\
        SCanNet$_{24}$ & 27.9 & 264.95 \\
        DEFO$_{24}$ & 26.02 & 401.09 \\
        CdSCNet$_{24}$ & 33.86 & 134.80 \\
        ChangeMamba$_{24}$ & 89.99 & 211.55 \\
        LSAFNet$_{24}$ & 27.86 & 521.92 \\
    \midrule
        Ours(Transformer-based)  & 86.75 & 169.79 \\
        Ours(Mamba-based)  & 86.32 & 183.68 \\
    \bottomrule
    \end{tabular}
    \label{tab:model_efficiency}
\end{table}

\subsubsection{Model efficiency}
\label{ssec:efficiency}


To further validate the efficiency of the proposed model, the model parameters (Params.) and floating-point operations per second (FLOPs) are reported in Table~\ref{tab:model_efficiency}, where the input is set to images of size 512 $\times$ 512 $\times$ 3 from the SECOND dataset. Compared with the Mamba-backbone-based ChangeMamba, our two variants not only contain fewer parameters and lower FLOPs, but also achieve superior accuracy. Specifically, the FLOPs of Transformer-based FoBa and Mamba-based FoBa are reduced to 80.25\% and 86.82\% of ChangeMamba, while the accuracy improvements of 1.71\% and 1.75\% on the SECOND dataset, respectively. And, compared with CNN-backbone-based methods such as DEFO and LSAFNet, our approach requires more parameters but exhibits substantially lower FLOPs, even comparable to the more lightweight SSCDl model. Overall, our proposed method demonstrates strong efficiency, while further reducing the parameter count remains a potential direction for future research.

\begin{figure*}
    \centering
    \includegraphics[width=0.7\textwidth]{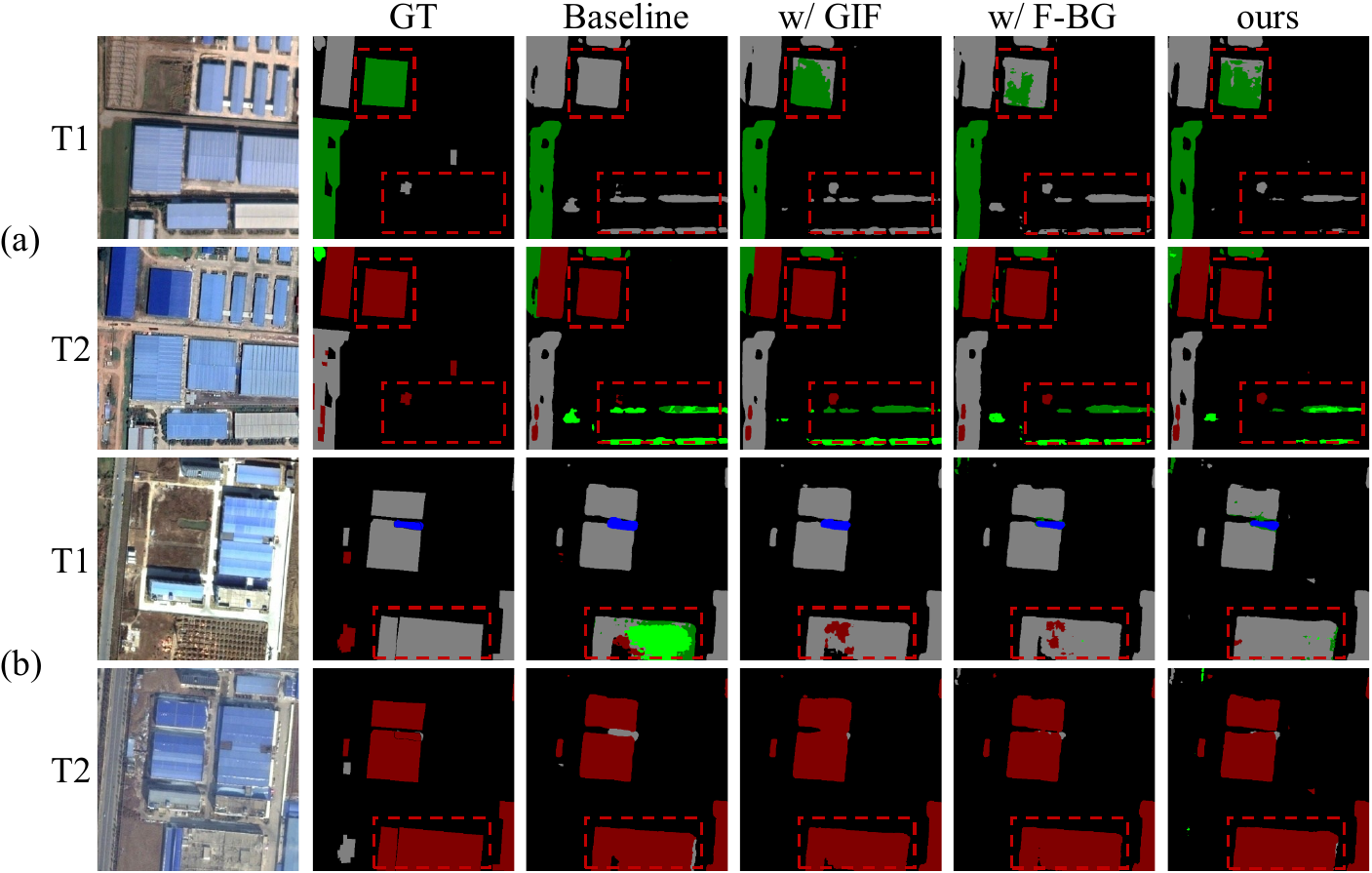}
    \caption{Visualization results of different variants on the SECOND dataset. W/ GIF denotes the baseline model equipped with GIF, while w/ F-BG indicates the baseline model with F-BG. The red bounding boxes highlight regions where our method yields superior results.}
    \label{fig:vis_different_compnents}
\end{figure*}

\subsection{Ablation studies}
\label{ssec:ablation}


In this subsection, we conduct a series of experiments on the SECOND and LevirSCD datasets to evaluate the impact of the key components on model performance, with detailed results presented in Table~\ref{tab:abalation_compnents}--\ref{tab:ablation_BG}.

\subsubsection{Effects of Different Components in FoBa}
\label{ssec: Different Components}


To verify the effectiveness of the key modules in the proposed FoBa framework, we design 8 ablation experiments on the SECOND and LevirSCD datasets. In addition, a variant obtained by removing the GIF and F-BG modules while keeping all other settings unchanged is adopted as the baseline for comparison. The results are reported in Table~\ref{tab:abalation_compnents}. It can be observed that the model performance consistently improves as the core modules are added, either individually or jointly. Compared with the baseline, the proposed method achieves gains of 0.66\%/0.85\% and 1.2\%/1.52\% on the SeK and $F_{scd}$, respectively. These notable improvements demonstrate that foreground-background co-guidance, bi-temporal feature interaction, and change-region consistency play critical roles in enhancing SCD performance. It is worth noting that the Mamba-based F-BG module achieves slightly better results than its Transformer-based counterpart. Therefore, all subsequent experiments and visualizations are reported using the Mamba-based variant. 

To further illustrate the advantages introduced by the key modules, we compare the visual results obtained by progressively integrating different components. As shown in the red-boxed mark of Fig.\ref{fig:vis_different_compnents} (a), the baseline completely misclassifies the area, whereas variants with the proposed modules significantly alleviate this issue. Moreover, in the Fig.\ref{fig:vis_different_compnents} (b), our model not only achieves the fewest misclassifications but also preserves the most complete change regions. The above visual results are consistent with the quantitative analysis, providing more intuitive evidence of the effectiveness of the proposed key modules.

\begin{table*}
    \centering
    \caption{Ablation study on Different Components.}
    \resizebox{0.9\textwidth}{!}{
    \begin{tabular}{l|c|c|c|c|c|c}
  \toprule
    \multicolumn{1}{c|}{} &
    \multicolumn{1}{c|}{} &
    \multicolumn{1}{c|}{} &
    \multicolumn{1}{c|}{} &
    \multicolumn{1}{c|}{} &
    \multicolumn{1}{c|}{\textbf{SECOND}}  &
    \multicolumn{1}{c}{\textbf{LevirSCD}} \\
    Model & GIF & \makecell{F-BG\\(Transformer-based)} & \makecell{F-BG\\(Mamba-based)} & \makecell{Consist.\\ Loss} & F$_{\text{scd}}$ / mIoU / Sek / OA & F$_{\text{scd}}$ / mIoU / Sek / OA \\
    \midrule
    Ours 
    & {$\times$}
    & {$\times$}
    & {$\times$}
    & {$\times$}
    & 56.72 / 73.88 / 23.95 / 85.77 
    & 39.75 / 76.38 / 18.33 / 89.42 \\
    \midrule
    Ours 
    & $\checkmark$
    & {$\times$}
    & {$\times$}
    & {$\times$}
    & 57.12 / 74.13 / 24.19 / 86.27  
    & 40.03 / 77.03 / 18.96 / 89.57 \\
    Ours 
    & $\times$
    & $\checkmark$
    & {$\times$}
    & {$\times$}
    & 57.01 / 74.38 / 24.33 / 86.02 
    & 40.06 / 76.87 / 18.86 / 89.48 \\
    Ours 
    & $\times$
    & {$\times$}
    & $\checkmark$
    & {$\times$}
    & 56.67 / 74.23 / 24.22 / 85.75 
    & 40.94 / 76.40 / 19.05 / 89.66 \\
    Ours 
    & $\checkmark$
    & $\checkmark$
    & {$\times$}
    & {$\times$}
    & 56.78 / 74.43 / 24.27 / 85.99  
    & 40.75 / 76.79 / 19.06 / 89.65 \\
    Ours
    & $\checkmark$
    & {$\times$}
    & $\checkmark$
    & {$\times$}
    & 57.07 / 74.48 / 24.47 / 86.02  
    & 41.29 / 76.68 / 19.40 / 89.71 \\
    \midrule
    Ours 
    & $\checkmark$
    & $\checkmark$
    & {$\times$}
    & {$\checkmark$}
    & 57.36 / 74.36 / 24.57 / 85.99
    & 40.82 / 76.99 / 19.33 / 89.28 \\
    Ours 
    & $\checkmark$
    & {$\times$}
    & $\checkmark$
    & {$\checkmark$}
    & 57.57 / 74.50 / 24.61 / 86.14
    & 41.27 / 76.82 / 19.53 / 89.76 \\
   \bottomrule
    \end{tabular}
    }
    \label{tab:abalation_compnents}
\end{table*}

\begin{table*}
    \centering
    \caption{Ablation study on different stages of GIF. And S stands for the stage.}
    \resizebox{0.8\textwidth}{!}{
    \begin{tabular}{l|c|c|c|c|c|c}
  \toprule
    \multicolumn{1}{c|}{} &
    \multicolumn{1}{c|}{} &
    \multicolumn{1}{c|}{} &
    \multicolumn{1}{c|}{} &
    \multicolumn{1}{c|}{} &
    \multicolumn{1}{c|}{\textbf{SECOND}}  &
    \multicolumn{1}{c}{\textbf{LevirSCD}} \\
    Model & S1 & S2 & S3 & S4 & F$_{\text{scd}}$ / mIoU / Sek / OA & F$_{\text{scd}}$ / mIoU / Sek / OA \\
    \midrule
    Ours w/o GIF
    & {$\times$}
    & {$\times$}
    & {$\times$}
    & {$\times$}
    & 56.67 / 74.23 / 24.22 / 85.75 
    & 40.94 / 76.40 / 19.05 / 89.66\\
    \midrule
    Ours 
    & $\checkmark$
    & {$\times$}
    & {$\times$}
    & {$\times$}
    & 57.12 / 74.25 / 24.32 / 86.22   
    & 40.15 / 77.22 / 19.10 / 89.40\\
    Ours 
    & $\checkmark$
    & $\checkmark$
    & {$\times$}
    & {$\times$}
    & 57.06 / 74.33 / 24.37 / 86.11   
    & 40.83 / 76.80 / 19.29 / 89.59\\
    Ours 
    & $\checkmark$
    & $\checkmark$
    & $\checkmark$
    & {$\times$}
    & 57.25 / 74.23 / 24.46 / 86.09  
    & 40.64 / 77.17 / 19.39 / 89.79\\
    Ours 
    & $\checkmark$
    & $\checkmark$
    & $\checkmark$
    & $\checkmark$
    & 57.57 / 74.50 / 24.61 / 86.14
    & 41.27 / 76.82 / 19.53 / 89.76 \\
   \bottomrule
    \end{tabular}
    }
    \label{tab:abalation_GIF}
\end{table*}

\subsubsection{Effects of different stages of GIF}
\label{ssec: different stages of GIF}


To further investigate the impact of GIF at different stages, we conduct experiments on the SECOND and LevirSCD datasets, with results reported in Table~\ref{tab:abalation_GIF}. Here, w/o GIF denotes the FoBa model without GIF, while S1-S4 correspond to adding GIF at different stages. As the GIF is progressively added at different stages, the model performance consistently improves trend. Specifically, compared with the variant without GIF, the proposed FoBa achieves gains of 0.39\%/0.9\% and 0.48\%/0.33\% on the key SCD metrics (SeK and $F_{scd}$) across the two datasets. This observation reflects the critical role of bi-temporal image interaction in SCD tasks.

\begin{table*}
    \centering
    \caption{Ablation study on different stages of F-BG. And D stands for the decoding stage.}
    \resizebox{0.8\textwidth}{!}{
    \begin{tabular}{l|c|c|c|c|c}
  \toprule
    \multicolumn{1}{c|}{} &
    \multicolumn{1}{c|}{} &
    \multicolumn{1}{c|}{} &
    \multicolumn{1}{c|}{} &
    \multicolumn{1}{c|}{\textbf{SECOND}}  &
    \multicolumn{1}{c}{\textbf{LevirSCD}} \\
    Model & D2 & D3 & D4 & F$_{\text{scd}}$ / mIoU / Sek / OA & F$_{\text{scd}}$ / mIoU / Sek / OA \\
    \midrule
    Ours w/o F-BG
    & {$\times$}
    & {$\times$}
    & {$\times$}
    & 57.12 / 74.13 / 24.19 / 86.27
    & 40.03 / 77.03 / 18.96 / 89.57\\
    \midrule
    Ours 
    & $\checkmark$
    & {$\times$}
    & {$\times$}
    & 57.17 / 74.38 / 24.34 / 86.11
    & 40.40 / 76.88 / 19.03 / 89.56\\
    Ours 
    & $\checkmark$
    & $\checkmark$
    & {$\times$}
    & 57.06 / 74.33 / 24.37 / 86.11
    & 40.69 / 76.97 / 19.23 / 89.44\\
    Ours 
    & $\checkmark$
    & $\checkmark$
    & $\checkmark$
    & 57.57 / 74.50 / 24.61 / 86.14 
    & 41.27 / 76.82 / 19.53 / 89.76\\
   \bottomrule
    \end{tabular}
    }
    \label{tab:abalation_F-BG}
\end{table*}

\subsubsection{Effects of different stages of F-BG}
\label{ssec: different stages of F-BG}


In addition, we also investigate the impact of introducing F-BG at different stages on model performance. Table~\ref{tab:abalation_F-BG} presents the experimental results on the SECOND and LevirSCD datasets. Here, w/o F-BG denotes the model without F-BG, while D2-D4 correspond to different decoding stages. Compared with the variant without F-BG, our model achieves improvements of 0.42\%/0.45\% and 0.57\%/1.24\% on the SeK and $F_{scd}$ metrics on two datasets, respectively. This demonstrates that foreground-background co-guidance can significantly enhance SCD performance. Notably, compared with the earlier stages, the D4 stage exhibits the most significant improvement, with SeK increasing by 0.24\% and 0.30\%. This is because the D4 stage can generate more refined guidance masks (i.e., masks with larger height and width), providing richer detailed information and thereby enhancing detection performance.

\begin{table*}
    \centering
    \caption{Ablation study on BG.}
    \resizebox{0.75\textwidth}{!}{
    \begin{tabular}{l|c|c|c|c}
  \toprule
    \multicolumn{1}{c|}{} &
    \multicolumn{1}{c|}{} &
    \multicolumn{1}{c|}{} &
    \multicolumn{1}{c|}{\textbf{SECOND}}  &
    \multicolumn{1}{c}{\textbf{LevirSCD}} \\
    Model & FG & BG & F$_{\text{scd}}$ / mIoU / Sek / OA & F$_{\text{scd}}$ / mIoU / Sek / OA \\
    \midrule
    Ours & $\times$ & $\times$
    & 57.12 / 74.13 / 24.19 / 86.27
    & 40.03 / 77.03 / 18.96 / 89.57\\
    Ours & $\checkmark$ & $\times$
    & 57.45 / 74.05 / 24.34 / 86.36
    & 41.00 / 76.70 / 19.26 / 89.68\\
    Ours & $\checkmark$ & $\checkmark$
    & 57.57 / 74.50 / 24.61 / 86.14
    & 41.27 / 76.82 / 19.53 / 89.76\\
   \bottomrule
    \end{tabular}
    }
    \label{tab:ablation_BG}
\end{table*}

\begin{figure*}
    \centering
    \includegraphics[width=0.9\textwidth]{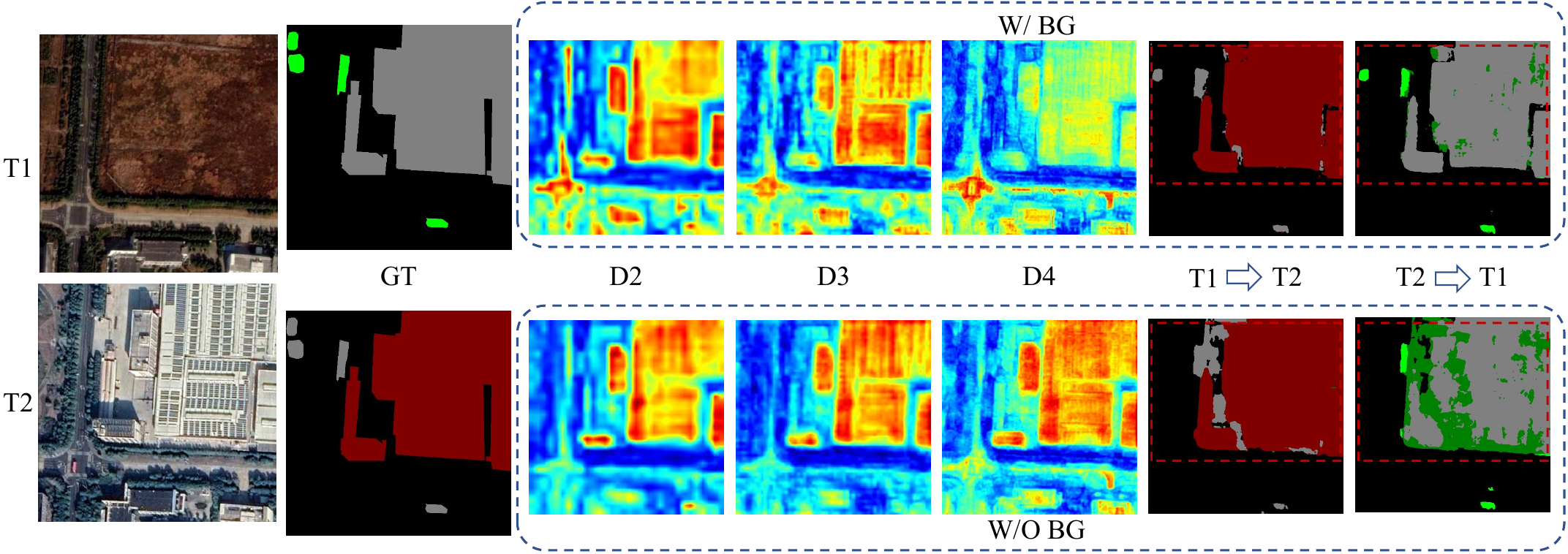}
    \caption{Visualization results of the decoder feature maps on the SECOND test set. W/O BG denotes the variant without BG, while w/ BG corresponds to our proposed method. \textcolor{red}{Red} denotes higher attention values, and \textcolor{blue}{blue} denotes lower values.}
    \label{fig:vis_bg}
\end{figure*}

\subsubsection{Effects of BG}
\label{ssec: Effects of BG}


To further investigate the impact of BG on model performance, we conduct experiments on the SECOND and LevirSCD datasets, with results presented in Table~\ref{tab:ablation_BG}. When only FG is added, the metrics reflecting semantic accuracy, SeK and $F_{scd}$, show improvements, while the mIoU metric, which reflects change-region accuracy, slightly decreases (from 77.03\% to 76.70\% on the LevirSCD dataset). This is because relying solely on foreground information overlooks smaller change regions, resulting in a reduction in mIoU. When both FG and BG are introduced, not only do the semantic accuracy metrics (SeK and $F_{scd}$) improve, but the change-region accuracy metric (mIoU) also shows certain gains, with an increase of approximately 0.5\% on the SECOND dataset. This highlights the importance of incorporating background information. 

To provide a more intuitive illustration of BG’s impact, we visualize the feature maps and prediction results at different decoding stages for the variants with and without BG, as shown in Fig.~\ref{fig:vis_bg}. Regarding the feature maps at different stages, compared with the w/o BG variant, which tends to overly focus on large change regions, the w/ BG variant achieves a more balanced attention across other regions (e.g. D4). In terms of SCD prediction results, the w/ BG variant exhibits more accurate semantic representations, particularly along the boundary regions, while also capturing finer change regions in the upper-left corner. These observations provide intuitive evidence that foreground–background co-guidance can substantially mitigate semantic ambiguity while enhancing the model’s sensitivity to subtle change regions.

\begin{table}
    \centering
    \caption{Effect of the different image encoders}
    \resizebox{0.45\textwidth}{!}{
    \begin{tabular}{c|c|c}
    \toprule
    \multicolumn{1}{c|}{} &
    \multicolumn{1}{c|}{Param.} &
    \multicolumn{1}{c}{\textbf{SECOND}} \\
    Backbone & (M) & F$_{\text{scd}}$ / mIoU / Sek / OA\\
        
    \midrule
    Tiny
    & 30.04 
    & 56.76 / 74.17 / 24.00 / 86.25  \\
    Small 
    & 49.63 
    & 57.25 / 74.08 / 24.11 / 86.53\\
    Base 
    & 86.32 
    & 57.57 / 74.50 / 24.61 / 86.14\\
    \bottomrule
    \end{tabular}
    }
    \label{tab:ablation_backbone}
\end{table}

\begin{table}
    \centering
    \caption{Effect of the different dims}
    \resizebox{0.35\textwidth}{!}{
    \begin{tabular}{c|c}
    \toprule
    \multicolumn{1}{c|}{} &
    \multicolumn{1}{c}{\textbf{LevirSCD}} \\
    Dims & F$_{\text{scd}}$ / mIoU / Sek / OA \\
        
    \midrule
    128
    & 41.27 / 76.82 / 19.53 / 89.76  \\
    256 
    & 40.57 / 77.13 / 19.34 / 89.64 \\
    512
    & 40.95 / 76.99 / 19.45 / 89.73\\
    768
    & 41.14 / 77.20 / 19.68 / 89.53\\
    1024
    & 40.44 / 77.49 / 19.49 / 89.61\\
    \bottomrule
    \end{tabular}
    }
    \label{tab:ablation_dims}
\end{table}

\begin{table}
    \centering
    \caption{Effect of coefficients of loss function}
    \resizebox{0.45\textwidth}{!}{
    \begin{tabular}{c|c|c|c|c}
    \toprule
    \multicolumn{1}{c|}{} &
    \multicolumn{1}{c|}{} &
    \multicolumn{1}{c|}{} &
    \multicolumn{1}{c|}{} &
    \multicolumn{1}{c}{\textbf{SECOND}} \\
    $\lambda_1$ & $\lambda_2$ & $\lambda_3$ & $\lambda_4$ & F$_{\text{scd}}$ / mIoU / Sek / OA \\
        
    \midrule
          1
          & 0.75
          & 0.5
          & 0.5
          & 57.57 / 74.50 / 24.61 / 86.14
          \\
          1
          & 1
          & 1
          & 1
          & 57.49 / 74.29 / 24.55 / 86.38
          \\        
          1
          & 0.75
          & 0.75
          & 0.75
          & 57.28 / 74.29 / 24.48 / 86.35
          \\
          1
          & 0.5
          & 0.5
          & 0.5
          & 57.15 / 74.24 / 24.33 / 86.34
          \\
    \bottomrule
    \end{tabular}
    }
    \label{tab:ablation_loss}
\end{table}

\subsection{Parameter Analysis}
\label{ssec:parameter}

\subsubsection{Different backbone} 

To investigate the impact of using different image encoders on the performance of the FoBa, we conduct experiments on the SECOND dataset, with results presented in Table~\ref{tab:ablation_backbone}. Here, tiny, small, and base denote different versions of VMamba. As the model version gradually increases, both SeK and $F_{scd}$ exhibit a consistent upward trend. Compared with the tiny version, the base version achieves a 0.61\% improvement in the SeK metric. Moreover, even with the most lightweight tiny version, whose parameter count is comparable to most existing sota models, the proposed method still achieves competitive performance. For instance, it attains a 0.87\% improvement over the relatively strong LSAFNet, further validating the effectiveness of the FoBa.

\subsubsection{Different dims in key module} 

To investigate the impact of different feature dimensions in the key modules (GIF and F-BG) on model performance, we conduct experiments on the LevirSCD dataset, with results presented in Table~\ref{tab:ablation_dims}. It can be observed that the SeK metric reaches its peak when the dimension is 768, and further increasing the dimension leads to a decline in performance. Since larger dimensions introduce a substantial number of parameters, for simplicity, we ultimately set the feature dimension of the FoBa to 128.

\subsubsection{Coefficients of Loss Function}

To investigate the impact of different loss function weights on model performance, we conduct experiments on the SECOND dataset, with results shown in Table~\ref{tab:ablation_loss}. It can be observed that when the weights of $L_{scd}+L_{cons}$, $L_{sample}$, and $L_f$ are equal (whether 0.5, 0.75, or 1), the model fails to achieve optimal performance, and the SeK metric decreases as the weights decrease. The best performance is achieved when $L_{scd}+L_{cons}$ is assigned a relatively larger weight. Ultimately, the weights of $L_{bcd}$, $L_{scd}+L_{cons}$, $L_{sample}$, and $L_f$ ($\lambda_1$-$\lambda_4$) are set to 1, 0.75, 0.5, and 0.5, respectively. 

\section{Conclusion}
\label{sec:conclusion}



In this paper, we introduce LevirSCD, a remote sensing semantic change detection dataset focused on the Beijing area. The dataset covers 16 common change categories, including paved road, low vegetation, and construction land, etc., encompassing 210 specific change types. It provides fine-grained category annotations (e.g., roads are divided into unpaved road and paved road) along with detailed object-level annotations. Additionally, the dataset provides semantic annotations for both T1 $\rightarrow$ T2 and T2 $\rightarrow$ T1. We expect that the proposed LevirSCD will facilitate further research in the field of remote sensing SCD.

Furthermore, we propose a novel semantic change detection method, FoBa, which effectively leverages clue from change information to achieve precise SCD. Specifically, we design Transformer-based and Mamba-based F-BG modules that, through a foreground-background co-guided strategy, enable the model to focus not only on the regions of interest but also to incorporate rich contextual background information. This strategy alleviates semantic ambiguity in complex scenes, particularly along object boundaries, and mitigates the issue of overly emphasizing large change regions while neglecting subtle changes. Moreover, considering the characteristics of bi-temporal feature interaction and change-region consistency in SCD tasks, we introduce the GIF module and an unchanged-region consistency loss to further enhance the model’s performance. Extensive ablation studies validate the effectiveness of the proposed modules. Meanwhile, experiments on two commonly used datasets, SECOND and JL1, as well as our proposed LevirSCD, demonstrate that our method achieves competitive performance.





\ifCLASSOPTIONcaptionsoff
  \newpage
\fi

{\small
\bibliographystyle{IEEEtran}
\bibliography{refs}
}


\end{document}